\pdfoutput=1

\documentclass[11pt]{article}

\usepackage[final]{acl}

\usepackage{times}
\usepackage{latexsym}

\usepackage[T1]{fontenc}

\usepackage[utf8]{inputenc}

\usepackage{microtype}

\usepackage{inconsolata}

\usepackage{hyperref}

\usepackage{graphicx}

\usepackage{amsmath}
\usepackage{amsfonts}
\usepackage{booktabs}
\usepackage{multirow}
\usepackage{algorithm}
\usepackage{algpseudocode}
\usepackage{caption}
\usepackage{subcaption}

\usepackage{pifont}
\usepackage{xspace}

\newcommand{\highlighthex}[3][1pt]{%
  {%
    \setlength{\fboxsep}{#1}
    \definecolor{_hlcolor}{HTML}{#2}
    \colorbox{_hlcolor}{\strut #3}
  }%
}

\newcommand{\cmark}[1]{%
  {\textcolor[HTML]{#1}{\ding{51}}}\xspace%
}

\newcommand{\xmark}[1]{%
  {\textcolor[HTML]{#1}{\ding{55}}}\xspace%
}

\title{Can LLMs Reliably Simulate Real Students' Abilities \\in Mathematics and Reading Comprehension?}

\author{
    KV Aditya Srivatsa\\
    \And
    Kaushal Kumar Maurya\\
    Mohamed bin Zayed University of Artificial Intelligence, Abu Dhabi, UAE \\
    {\tt \{vaibhav.kuchibhotla, kaushal.maurya, ekaterina.kochmar\}@mbzuai.ac.ae} \\
    \And
    Ekaterina Kochmar
}

\begin{document}
\maketitle
\begin{abstract}

Large Language Models (LLMs) are increasingly used as \textit{proxy students} in the development of Intelligent Tutoring Systems (ITSs) and in piloting test questions. However, \textit{to what extent these proxy students accurately emulate the behavior and characteristics of real students remains an open question.}  To investigate this, we collected a dataset of 489 items from the National Assessment of Educational Progress (NAEP), covering mathematics and reading comprehension in grades 4, 8, and 12. We then apply an \textit{Item Response Theory (IRT)} model to position 11 diverse and state-of-the-art LLMs on the same ability scale as real student populations. Our findings reveal that, without guidance, strong general-purpose models consistently outperform the average student at every grade, while weaker or domain-mismatched models may align incidentally. Using grade-enforcement prompts changes models' performance, but whether they align with the average grade-level student remains highly model- and prompt-specific: no evaluated model–prompt pair fits the bill across subjects and grades, underscoring the need for new training and evaluation strategies. We conclude by providing guidelines for the selection of viable proxies based on our findings.\footnote{All related code and data is available at \href{https://github.com/kvadityasrivatsa/IRT-for-LLMs-as-Students}{https://github.com/kvadityasrivatsa/IRT-for-LLMs-as-Students}}
\end{abstract}

\section{Introduction}
\label{sec:introduction}

\begin{figure}[!ht]
    \centering
    \includegraphics[width=0.9\linewidth]{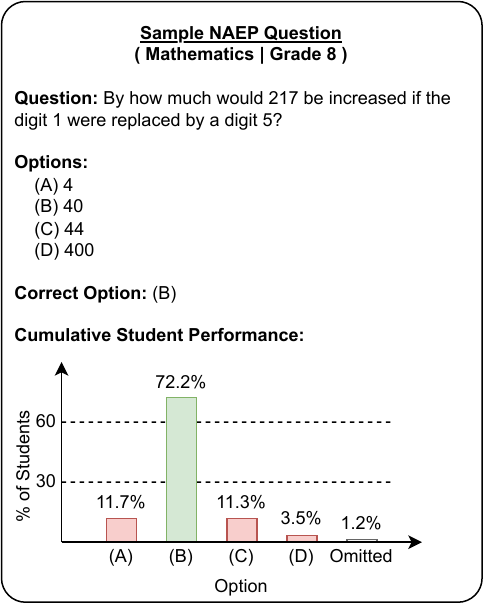}
    \caption{Sample NAEP question from grade 8 mathematics.}
    \label{fig:naep-example-g8-math}
\end{figure}

Large language models (LLMs) are capable of generating fluent and coherent text and excelling at many complex tasks \cite{chang2024survey, zhao2024explainability}. Their rise offers new opportunities for educational technology, notably in (i) intelligent tutoring systems (ITS; \citealp{wang2024large}) and (ii) \emph{piloting assessments} before they go live \cite{Liu2025, Rehse2024}. ITS provides targeted feedback and adaptive instruction, while reliable assessments track learning without bias. Yet both require understanding how real students would interact with them, which is extremely challenging to verify.

Ideally, tutors and test forms should be vetted on representative student samples across skill levels. This, however, is resource-intensive, especially in regions already short on teachers and infrastructure \cite{unesco2023tech, woolf2013developing}. Teacher-led evaluations (e.g., \citealp{macina-etal-2023-mathdial}) and static logs similarly fail to scale or capture the dynamics of new items and adaptive strategies \cite{belz-etal-2023-non, ed2023ai}. These constraints motivate alternatives that enable rigorous, equitable evaluation at scale.

An emerging approach is for \emph{simulate} students with LLMs \cite{WhartonLLMClass2024,sonkar2024llm}. Proxy models can be conditioned on attributes such as grade level, offering fast, repeatable tests of tutor features or item quality. However, current evaluations are based on an expert judgment of plausibility \cite{macina-etal-2023-mathdial}, leaving open the question of how closely such proxies match real student performance.
Similarly, in psychometrics, LLMs have been used as \emph{synthetic examinees}: e.g., \citet{Liu2025} show that {\tt GPT-3.5/4} answer sets yield item statistics that mirror a 50-student pilot, and \citet{Rehse2024} demonstrate that ChatGPT can pre-flag weak or biased items. However, these studies treat LLMs as single test-takers and do not look into whether persona prompts can tie their abilities to specific grade bands.

\textbf{Our approach:} We apply IRT \cite{baker2001basics} to measure how 11 diverse LLMs and real students perform on the same grade-level questions. Using data from the National Assessment of Educational Progress (NAEP) \cite{nces2022naep} for mathematics and reading in grades 4, 8, 12, we evaluate whether the LLM responses (both \emph{generic} and \emph{grade-conditioned}) align with authentic student response patterns. 
Specifically, we address the following research questions: \textbf{RQ1} -- Under standard prompting, how do LLMs compare with real students across grades and subjects? \textbf{RQ2} -- When asked to act as an average student in a given grade: How does LLM performance change? (\textbf{RQ2.1}) Does the shift match real grade-level patterns? (\textbf{RQ2.2}) 

The main contributions of our work are as follows:
\begin{itemize}
    \item We compile and release a dataset\footnote{\href{https://github.com/kvadityasrivatsa/IRT-for-LLMs-as-Students}{https://github.com/kvadityasrivatsa/IRT-for-LLMs-as-Students}} sourced from NAEP of real student responses to subject-specific, grade-targeted questions, covering two subjects (mathematics and reading assessment) and three grade levels (4, 8 and 12). 
    \item We adapt Item Response Theory (IRT) to assess the alignment between LLM-generated responses and actual student performance patterns.
    \item We conduct an evaluation of 11 diverse LLMs, examining how well they approximate student responses under both \textit{unenforced} (generic) and \textit{grade-enforced} prompts.
\end{itemize}

\section{Related Work}
\label{sec:related-work}

\paragraph{Simulated Students in Intelligent Tutoring Systems}
Early simulated-student work relied on production rule \emph{apprentices} that learn from worked examples and then reproduce step-level behavior inside an ITS.  
The \textit{SimStudent} / {\em Apprentice-Learner} family shows that such models can generate realistic error types and serve as policy learners to hint \citep{Matsuda2023,MacLellan2024}.  
More recent studies graft LLMs onto this pipeline: it has been shown that {\tt GPT-4} ``think-aloud’’ traces improve bug-library discovery and fine-grained skill tagging, while LLM agents at dialogue level can populate entire synthetic classroom cohorts \citep{WhartonLLMClass2024}.  
These approaches demonstrate that generative text can complement symbolic learner models, yet they rarely test whether the \emph{ability} distribution of synthetic learners matches that of real students -- a gap we address through our analysis.

\paragraph{LLM-Generated Responses for Item Calibration}
Psychometric studies have begun to treat LLM outputs as \emph{synthetic examinee responses}.
\citet{Liu2025} show that the {\tt GPT-3.5/4} answer sets yield 3PL item statistics that match a 50-student baseline, reducing the pretest costs.
\citet{Rehse2024} use ChatGPT to filter out low information or biased items.
\citet{heyueya2024psychometricalignmentcapturinghuman} further adapt IRT to align LLM and human response patterns, while \citet{zelikman-etal-2023-generating} simulate K-12 students.
However, these works produce only aggregate correlations; they do not examine whether an LLM’s \emph{latent ability} aligns with a particular grade band or whether persona prompts shift that ability in predictable ways.  
Our work closes this gap with an IRT model that maps LLM performance to grade-level performance.

\paragraph{Persona-Conditioned Prompting and Alignment}
Prompting a model with an explicit role (e.g., ``\textit{You are a 4\textsuperscript{th}-grade student}’’) can change both reasoning depth and surface style.  
\citet{benedetto-etal-2024-using} find that a one-sentence student-level prompt lets {\tt GPT-4} imitate weak, average, and strong test-takers across subjects, although adherence to the target level is uneven.  
Broader evaluations such as CharacterEval \citep{Tu2024} measure persona consistency in dialogues, while \citet{KimJung2024} show that role prompts can either help or hurt accuracy depending on task characteristics.  
None of these efforts connect persona adherence to \emph{quantitative} grade-level ability estimates, nor do they compare default and persona-conditioned ability curves within a unified IRT framework.

\medskip
Together, these strands indicate that (i) LLMs are already employed as simulated students and psychometric stand-ins, and (ii) persona prompts shift model behavior without a principled link to grade-level metrics.  
Our study unifies the two directions by applying an IRT model to quantify how default and persona-conditioned LLM outputs align with average student performance at grades 4, 8, and 12.

\section{NAEP Data}
\label{sec:dataset}

\subsection{Source and Composition}

We prepared our dataset using publicly available items and student response data from the National Assessment of Educational Progress (NAEP) \cite{nces2022naep},\footnote{\url{https://nces.ed.gov/nationsreportcard/}} a large-scale assessment program administered by the National Center for Education Statistics (NCES). NAEP periodically assesses student achievement across the United States in key subject areas, including mathematics and reading. These assessments are conducted in grades 4, 8, and 12, offering a cross-sectional perspective on student proficiency throughout K–12 education.

\subsection{Coverage and Educational Context}

We source questions from both the \textit{mathematics} and \textit{reading comprehension} assessments at the three grade levels, capturing a broad spectrum of student performance and cognitive development throughout different educational stages. We focus on these two subjects for two reasons: (1) numeracy and literacy are considered fundamental skills (e.g., \citealp{williams2003skills}); and (2) NAEP data cover three grades for these subjects, while many other subjects only cover one or two grades. Math questions span topics such as measurement, algebra, geometry, and probability and statistics, with overall difficulty scaling with grade level. Reading comprehension items are based on passages whose average length increases with grade. The corresponding questions shift from direct factual queries in lower grades to those requiring interpretation and reflection at higher levels.

Each record contains the original question, multiple choice options, the correct annotated answer, and anonymized aggregate response patterns. For each item, the dataset reports the percentage of students who selected each option or omitted the question. Figures \ref{fig:naep-example-g8-math} and \ref{fig:naep-example-g12-reading} show representative examples from the grade-8 mathematics and grade-12 reading subsets, respectively.

Since NAEP is a continually administered assessment, this dataset can be periodically updated with newly released items. This makes it a dynamic resource that can evolve along with changes in educational standards and student performance distributions, offering long-term utility for evaluating automated student proxies and similar tasks.

\subsection{Preprocessing and Filtering Criteria}

NAEP assessments encompass a variety of question types, modalities, and response formats. Given that this is a preliminary effort to develop a quantitative and interpretable framework for aligning LLM performance with real student behavior, the inclusion of diverse modalities can introduce confounding factors that obscure analysis. For example, items that involve images, diagrams, or tables introduce the additional variable of visual comprehension, making it difficult to isolate language understanding as the primary factor in model performance. Similarly, free-form responses present evaluation challenges: gold-standard answers are often limited in number and may not capture the full range of acceptable responses. Assessing these reliably often requires expert judgment, which undermines the feasibility of scalable, LLM-based evaluation.

In contrast, multiple choice questions offer clearly defined answer sets, enabling a more straightforward and objective evaluation, which is crucial for both quantitative benchmarking and interpretability via Item Response Theory (IRT). Consequently, we apply two main filtering criteria when constructing our dataset.

\begin{itemize} 
    \item \textit{Text-only content:} We exclude any items that involve diagrams, tables, or multimedia elements, retaining only questions and instructions presented in text. \item \textit{Multiple-choice format:} We include only multiple choice questions (MCQs), which support standardized evaluation and facilitate downstream processing, such as answer extraction and IRT-based analysis. 
\end{itemize}

\begin{figure}[!ht]
    \centering
    \includegraphics[width=\columnwidth]{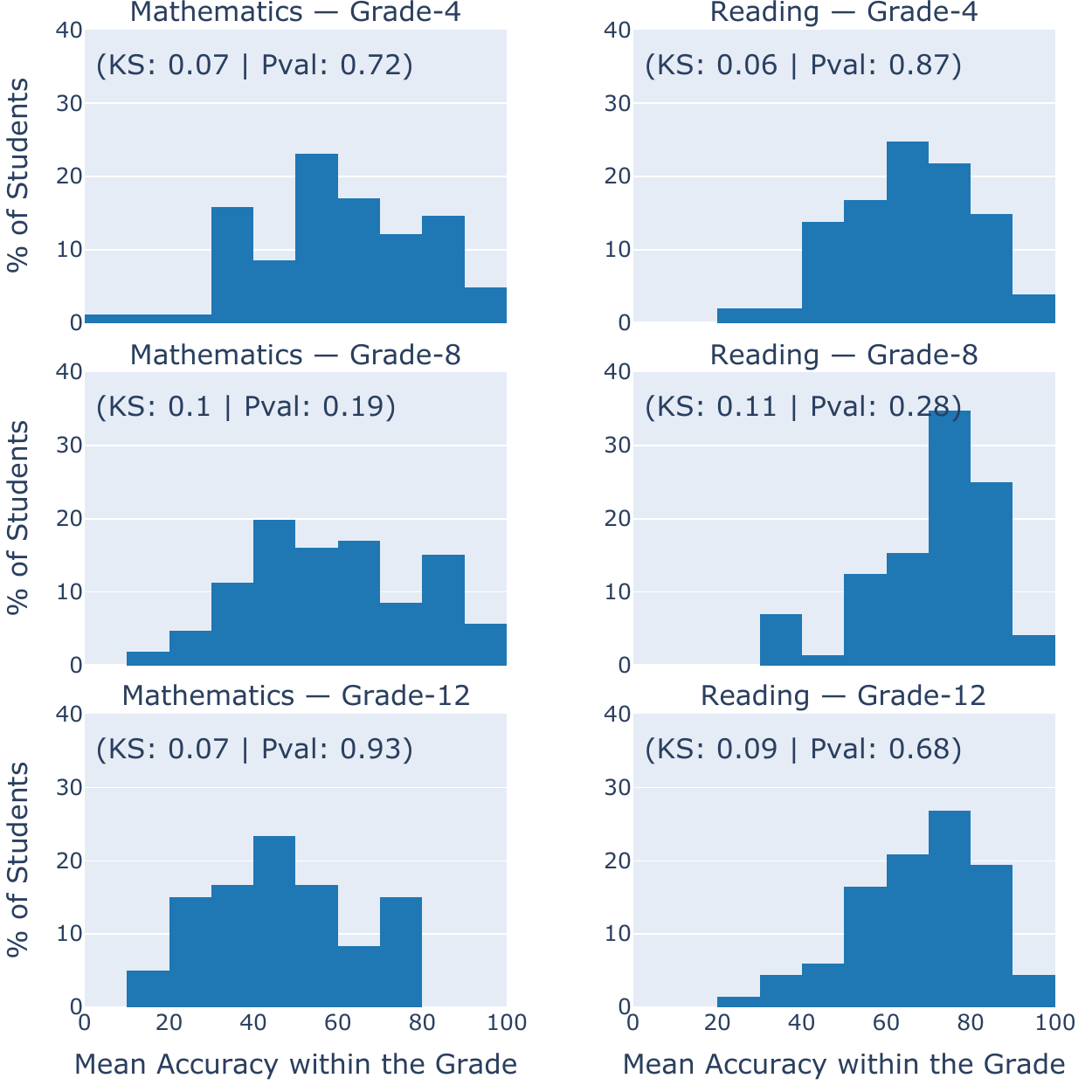}
    \caption{Distribution of question-level accuracy in NAEP assessments across grades and subjects. KS statistics and corresponding p-values are reported to assess normality; distributions with p > 0.05 are considered consistent with a normal distribution.}
    \label{fig:data-distr}
\end{figure}

After filtering, our final dataset consists of 489 multiple-choice items in English: 249 from mathematics and 240 from reading. Table~\ref{tab:data-stats} summarizes key statistics from the dataset. Figure \ref{fig:data-distr} shows the distribution of questions answered by students with varying accuracy for each subject and grade. We calculate the Kolmogorov-Smirnov (KS) statistic for each distribution to test for normality, and all subsets sufficiently (p-value > 0.05) follow a normal distribution. 

\section{Proposed Methodology}
\label{sec:methodology}

Our goal is to evaluate LLMs alongside human students recorded in the NAEP dataset by estimating their answering ability on a shared scale. To do this, we draw on Item Response Theory (IRT; \citet{baker2001basics}), a well-established framework in educational measurement. IRT enables us to jointly model the ability of test takers and the difficulty of individual test items using probabilistic principles.

\subsection{Estimating Student (LLM) Ability}

We begin with the Rasch model \cite{rasch1960probabilistic}, a widely used and interpretable form of IRT. It assumes that the probability of a correct response depends only on the difference between a participant's ability and an item's difficulty. This model uses a single parameter per item, that is, difficulty \( b_i \), and one ability parameter \( \theta_i \) per participant.

The Rasch model defines the probability that participant \( i \) correctly answers item \( j \) as:

\begin{equation}
P(X_{ij} = 1) = \frac{e^{\theta_i - b_j}}{1 + e^{\theta_i - b_j}},
\label{eq:rasch}
\end{equation}

\noindent where \( \theta_i \in \mathbb{R} \) is the ability of the participant \( i \), \( b_j \in \mathbb{R} \) is the difficulty of item \( j \), and $\mathbb{R}$ is the common scale of difficulty or ability.

To estimate \( b_j \), we use the empirical proportion \( p_j \) of correct responses for each item in the population. A simple approximation is:

\begin{equation}
b_j \approx \log\left( \frac{1 - p_j}{p_j} \right),
\label{eq:alpha}
\end{equation}

\noindent which reflects that the harder items (with lower \( p_j \)) have higher difficulty values \cite{BondFox2015}. Once the item difficulties are known, the ability of each participant \( \theta_i \) can be estimated using marginal maximum likelihood or Bayesian inference based on their response pattern.

\subsection{Grade-Alignment Prompting}

Our first research question (RQ1) investigates how an LLM's problem-solving ability compares to that of average students at different grade levels, specifically grades 4, 8 and 12, based on the NAEP dataset. To measure this, we begin with a minimal zero-shot prompt, which we refer to as \textbf{\textsc{Unenforced}} (see Appendix Figures \ref{fig:math-minimal-unenforced} and \ref{fig:reading-minimal-unenforced} for exact prompt templates). This prompt simply presents the question to the model without any added instructions or persona guidance.

Our second research question (RQ2) explores whether LLMs can align their responses with the answering patterns and performance levels of students in specific grades. To probe this, we design a set of increasingly guided zero-shot prompts that aim to steer the model toward grade-level reasoning.

\begin{enumerate}
    \itemsep-0em
    \item \textbf{\textsc{GradeEnforcedMinimal}}: Identical to the {\tt Unenforced} prompt, but with the added instruction that the model should act as an average student from a specific grade (4, 8, or 12). The exact prompts are presented in Appendix Figures \ref{fig:math-minimal-enforced} and \ref{fig:reading-minimal-enforced}.
    
    \item \textbf{\textsc{GradeEnforcedBasicCoT}}: Builds on the minimal version by prompting the model to consider what an average student at the specified grade would likely choose and why. This prompt encourages brief, grade-aware reasoning and reflects the student’s typical reasoning ability and common error patterns. See Figures \ref{fig:math-basic-cot-enforced} and \ref{fig:reading-basic-cot-enforced} in the Appendix for the exact prompts.
    
    \item \textbf{\textsc{GradeEnforcedFullCoT}}: Adds further scaffolding by dividing the reasoning process into two steps. First, the model is instructed to reflect on whether an average student at the given grade level would be likely to answer the question correctly. Second, based on that reflection, the model either justifies a correct answer or, if the student is unlikely to succeed, selects and explains the most plausible incorrect answer. See Figures \ref{fig:math-full-cot-enforced} and \ref{fig:reading-full-cot-enforced} in the Appendix for the exact prompts.
\end{enumerate}

The design of the \textsc{GradeEnforcedBasicCoT} and \textsc{GradeEnforcedFullCoT} prompts is inspired by \citet{benedetto-etal-2024-using}, who developed similar prompts to simulate student reasoning across skill levels on exam-style questions of varying difficulty. Their work informed our decision to incorporate reasoning about the ability of a student and the likelihood of error into our prompt design.

Our aim is not to claim these are optimal prompts or to exhaustively search for the best possible formulations. Instead, we adopt straightforward, representative prompting strategies aligned with popular practices to focus our investigation on whether such methods meaningfully promote grade-level alignment in model behavior. This may limit the scope of our findings, but it allows us to isolate and evaluate the effects of targeted prompting on grade-sensitive reasoning.

\section{Experimental Setup}
\label{sec:experiments}

\subsection{Task Setup}

We design our experiments based on the framework described in Section~\ref{sec:methodology}. The evaluation is conducted in two phases:

\begin{enumerate}
    \item \textbf{Problem-Solving:} LLMs answer questions in a standard problem solving setting, without specific instructions on how to mimic human behavior. Their performance is compared to that of average students in different grade levels.
    
    \item \textbf{Grade-Level Mimicking:} LLMs are explicitly instructed to emulate an average student of a specific grade level and respond as such.
\end{enumerate}

In both phases, we apply a Rasch model to assess performance. Each question is treated as an individual item $j$, and each LLM is treated as an in-distribution test-taker $i$. The binary response of LLM $i$ to the question $j$ is represented as $s_{ij} \in \{0,1\}$, where $s_{ij} = 1$ indicates a correct answer.

\begin{table}[!ht]
\centering
\resizebox{\columnwidth}{!}{%
\begin{tabular}{@{}lccccc@{}}
\toprule[1.5pt]
\multicolumn{1}{c}{\multirow{2}{*}{\textbf{LLM}}} & \multirow{2}{*}{\textbf{\begin{tabular}[c]{@{}c@{}}Open\\ Source?\end{tabular}}} & \multicolumn{1}{c}{\multirow{2}{*}{\textbf{\begin{tabular}[c]{@{}c@{}}Parameter\\ Count\end{tabular}}}} & \multirow{2}{*}{\textbf{\begin{tabular}[c]{@{}c@{}}Fine-\\ Tuned?\end{tabular}}} & \multicolumn{2}{c}{\textbf{Benchmark Scores}} \\ \cmidrule(l){5-6} 
\multicolumn{1}{c}{} &  & \multicolumn{1}{c}{} &  & {\tt GSM8K (\%)} & {\tt MMLU (\%)} \\ \midrule[1pt]
{\tt LLaMA2-13B} \cite{LLaMA2} & \cmark{056517} & 13B & \xmark{BF1029} & 28.7 & 54.8 \\ \midrule
{\tt LLaMA2-70B} \cite{LLaMA2} & \cmark{056517} & 70B & \xmark{BF1029} & 56.8 & 68.9 \\ \midrule
{\tt LLaMA3.1-8B} \cite{LLaMA2} & \cmark{056517} & 8B & \xmark{BF1029} & 84.5 & 73.0 \\ \midrule
{\tt LLaMA3.1-70B} & \cmark{056517} & 70B & \xmark{BF1029} & 95.1 & 86.0 \\ \midrule
{\tt Mistral-7B} \cite{mistral} & \cmark{056517} & 7B & \xmark{BF1029} & 52.1 & 60.1 \\ \midrule
{\tt Qwen2.5-7B} \cite{qwen25} & \cmark{056517} & 7B & \xmark{BF1029} & 85.4 & 74.2 \\ \midrule
{\tt Qwen2.5-Math} \cite{qwen25} & \cmark{056517} & 7B & \cmark{056517} & 91.6 & 67.8 \\ \midrule
{\tt GPT-3.5-Turbo} \cite{openai2023gpt3_5_turbo} & \xmark{BF1029} & {\tt --} & \xmark{BF1029} & 57.1 & 70.0 \\ \midrule
{\tt o3-Mini} \cite{openai2025o3_mini} & \xmark{BF1029} & {\tt --} & \xmark{BF1029} & 89.9 & 85.2 \\ \midrule
{\tt SocraticLM} \cite{socraticlm_2024} & \cmark{056517} & 7B & \cmark{056517} & 60.6 & -- \\ \midrule
{\tt LearnLM-1.5-Pro} & \xmark{BF1029} & {\tt --} & \cmark{056517} & -- & -- \\ 
\cite{learnlm_15_pro} & & & & & \\
\bottomrule[1pt]
\end{tabular}%
}
\caption{List of LLMs evaluated in our study, along with key descriptors about each model, i.e., open source availability, parameter size, whether the model is fine-tuned (as opposed to pretrained or instruction-tuned), and scores on reasoning and comprehension benchmarks GSM8K and MMLU (we omit scores that have not been released publicly by the respective model's paper or technical report).}
\label{tab:model-list}
\end{table}

\subsection{Models}
We select a diverse set of 11 LLMs (see Table \ref{tab:model-list}) to ensure broad coverage across access types (open vs. closed), model sizes and training paradigms (pretrained vs. domain-finetuned). Our goal is to capture a range of capabilities relevant to reasoning and comprehension, as reflected in benchmarks like {\tt GSM8K}~\cite{cobbe2021training} and {\tt MMLU}~\cite{hendrycks2020measuring}. We include both general-purpose models and those finetuned to specific domains. {\tt GPT-3.5-Turbo} is included based on \citet{benedetto-etal-2024-using}, who suggest that it can adapt responses to the levels of student ability instructed. {\tt SocraticLM} and {\tt LearnLM-1.5-Pro} are fine-tuned on pedagogical data; therefore, they might have more accurate insights into the performance of students at different grade levels.

\subsection{Evaluation}

\paragraph{Measuring Problem-Solving Correctness}  
All problems in our dataset are multiple choice questions (MCQs), which simplifies the evaluation of correctness: \textit{a model's response is considered correct if the selected option matches the correct answer provided with the dataset.} This binary distinction between correct and incorrect responses makes the data well-suited for dichotomous (i.e., the answer can only be correct or incorrect) Item Response Theory (IRT) models. We found that model responses vary in structure and require a unified follow-up prompt to extract the predicted choice from each model response (see Figure \ref{fig:option-extraction} in the Appendix).

\paragraph{Estimating Grade-Level Alignment}  
To address both research questions, we estimate how closely a model's performance aligns with that of an average student at a given grade level. 
To pin the origin and unit of the Rasch ability scale during marginal maximum-likelihood estimation, we follow the standard convention of treating examinee abilities as standard-normal, $\theta \sim \mathcal{N}(0,1)$. Although not theoretically necessary, \citet{EmbretsonReise2000} note that this assumption is a reasonable way to identify the latent trait because it fixes the zero point and variance without constraining the shape of the data.

Therefore, the average student has an ability parameter of zero ($\theta_{\text{avg}} = 0$).

The estimated ability parameter $\theta_i$ for a model $i$ can be interpreted in relation to this benchmark. \textit{The closer $\theta_i$ is to zero, the more the model's performance aligns with that of the average student.} We can also express this alignment using the percentile rank, computed via the cumulative distribution function (CDF) of the standard normal distribution, denoted by \( \Phi(\theta) \):

\begin{equation}
    \text{Percentile Rank} = \Phi(\theta) \times 100
\end{equation}

A percentile rank of 50 corresponds to the average student. Higher percentile ranks indicate higher levels of ability relative to the population. We use percentile rank as our main metric to measure LLM or student ability, as it has a fixed range (0-100 and is centered at 50), which allows for easier comparison of LLM alignment across subjects and grade levels.

\begin{table*}[!ht]
\centering
\resizebox{\textwidth}{!}{%
\begin{tabular}{@{}lccccccccc|ccccccccc@{}}
\toprule[1.5pt]
 & \multicolumn{9}{c|}{\textbf{Mathematics}} & \multicolumn{9}{c}{\textbf{Reading}} \\ \cmidrule(l){2-19} 
 & \multicolumn{3}{c|}{\textbf{Question Grade 4}} & \multicolumn{3}{c|}{\textbf{Question Grade 8}} & \multicolumn{3}{c|}{\textbf{Question Grade 12}} & \multicolumn{3}{c|}{\textbf{Question Grade 4}} & \multicolumn{3}{c|}{\textbf{Question Grade 8}} & \multicolumn{3}{c}{\textbf{Question Grade 12}} \\ \cmidrule(l){2-19} 
\multirow{-3}{*}{\textbf{LLM}} & \textbf{$P_{U}$} & \textbf{$P_{E}$} & \multicolumn{1}{c|}{\textbf{$\Delta$}} & \textbf{$P_{U}$} & \textbf{$P_{E}$} & \multicolumn{1}{c|}{\textbf{$\Delta$}} & \textbf{$P_{U}$} & \textbf{$P_{E}$} & \textbf{$\Delta$} & \textbf{$P_{U}$} & \textbf{$P_{E}$} & \multicolumn{1}{c|}{\textbf{$\Delta$}} & \textbf{$P_{U}$} & \textbf{$P_{E}$} & \multicolumn{1}{c|}{\textbf{$\Delta$}} & \textbf{$P_{U}$} & \textbf{$P_{E}$} & \textbf{$\Delta$} \\ \midrule[1.5pt]
{\tt LLaMA2-13B} & \cellcolor[HTML]{75A6F7}{\color[HTML]{FFFFFF} \textbf{63.7}} & \cellcolor[HTML]{75C38A}66.1 & \multicolumn{1}{c|}{\cellcolor[HTML]{F3F2EE}+2.4} & \cellcolor[HTML]{4B8BF4}{\color[HTML]{FFFFFF} \textbf{52.6}} & \cellcolor[HTML]{37A955}{\color[HTML]{FFFFFF} \textbf{50.8}} & \multicolumn{1}{c|}{\cellcolor[HTML]{F4F1EE}-1.8} & \cellcolor[HTML]{4889F5}{\color[HTML]{FFFFFF} \textbf{48.6}} & \cellcolor[HTML]{77C48B}66.5 & \cellcolor[HTML]{F4EAC9}+17.9 & \cellcolor[HTML]{FDFEFE}99.7 & \cellcolor[HTML]{ECF7EF}95.5 & \multicolumn{1}{c|}{\cellcolor[HTML]{F4F0E9}-4.2} & \cellcolor[HTML]{EBF2FD}94.9 & \cellcolor[HTML]{CCE9D4}87.6 & \multicolumn{1}{c|}{\cellcolor[HTML]{F4EFE1}-7.3} & \cellcolor[HTML]{B6D0FA}80.9 & \cellcolor[HTML]{57B770}58.7 & \cellcolor[HTML]{F5E6BD}-22.3 \\ \midrule
{\tt LLaMA2-70B} & \cellcolor[HTML]{C8DBFB}85.5 & \cellcolor[HTML]{77C58C}33.5 & \multicolumn{1}{c|}{\cellcolor[HTML]{F8D676}-52.0} & \cellcolor[HTML]{A5C5FA}23.9 & \cellcolor[HTML]{7FC993}31.6 & \multicolumn{1}{c|}{\cellcolor[HTML]{F3EFE1}+7.6} & \cellcolor[HTML]{5F98F6}42.4 & \cellcolor[HTML]{53B56D}57.8 & \cellcolor[HTML]{F4EBCF}+15.4 & \cellcolor[HTML]{D3E3FC}88.6 & \cellcolor[HTML]{ACDBB9}79.8 & \multicolumn{1}{c|}{\cellcolor[HTML]{F4EEDD}-8.8} & \cellcolor[HTML]{96BBF8}72.3 & \cellcolor[HTML]{81C994}69.1 & \multicolumn{1}{c|}{\cellcolor[HTML]{F4F1EB}-3.2} & \cellcolor[HTML]{93B9F8}71.6 & \cellcolor[HTML]{57B770}58.7 & \cellcolor[HTML]{F5EBD4}-12.9 \\ \midrule
{\tt LLaMA3.1-8B} & \cellcolor[HTML]{F2F7FE}96.8 & \cellcolor[HTML]{C4E5CD}85.5 & \multicolumn{1}{c|}{\cellcolor[HTML]{F4ECD8}-11.3} & \cellcolor[HTML]{C6DAFB}85.2 & \cellcolor[HTML]{5CB975}60.0 & \multicolumn{1}{c|}{\cellcolor[HTML]{F6E5B6}-25.2} & \cellcolor[HTML]{B1CCFA}79.5 & \cellcolor[HTML]{82C995}69.3 & \cellcolor[HTML]{F4EDDA}-10.2 & \cellcolor[HTML]{FDFEFE}99.7 & \cellcolor[HTML]{A5D8B2}77.9 & \multicolumn{1}{c|}{\cellcolor[HTML]{F5E7BE}-21.8} & \cellcolor[HTML]{F2F6FE}96.7 & \cellcolor[HTML]{E1F2E6}92.7 & \multicolumn{1}{c|}{\cellcolor[HTML]{F4F0E9}-4.0} & \cellcolor[HTML]{CCDEFC}86.7 & \cellcolor[HTML]{BDE2C7}83.9 & \cellcolor[HTML]{F4F1EC}-2.8 \\ \midrule
{\tt LLaMA3.1-70B} & \cellcolor[HTML]{FDFEFE}99.6 & \cellcolor[HTML]{F5FAF6}97.6 & \multicolumn{1}{c|}{\cellcolor[HTML]{F4F1EE}-2.0} & \cellcolor[HTML]{FAFCFE}98.9 & \cellcolor[HTML]{F7FBF8}98.0 & \multicolumn{1}{c|}{\cellcolor[HTML]{F4F2F0}-0.9} & \cellcolor[HTML]{F0F5FE}96.1 & \cellcolor[HTML]{F2F9F4}97.0 & \cellcolor[HTML]{F3F3F1}+0.9 & \cellcolor[HTML]{FEFEFE}99.9 & \cellcolor[HTML]{FEFEFE}99.9 & \multicolumn{1}{c|}{\cellcolor[HTML]{F3F3F3}0.0} & \cellcolor[HTML]{F2F6FE}96.7 & \cellcolor[HTML]{E1F2E6}92.7 & \multicolumn{1}{c|}{\cellcolor[HTML]{F4F0E9}-4.0} & \cellcolor[HTML]{C2D7FB}83.9 & \cellcolor[HTML]{B1DDBD}80.9 & \cellcolor[HTML]{F4F1EB}-3.0 \\ \midrule
{\tt Mistral-7B} & \cellcolor[HTML]{75A6F7}{\color[HTML]{FFFFFF} \textbf{63.7}} & \cellcolor[HTML]{58B771}58.9 & \multicolumn{1}{c|}{\cellcolor[HTML]{F4F0E7}-4.8} & \cellcolor[HTML]{5B95F6}43.5 & \cellcolor[HTML]{39AA57}49.0 & \multicolumn{1}{c|}{\cellcolor[HTML]{F3F1E7}+5.4} & \cellcolor[HTML]{75A6F7}63.7 & \cellcolor[HTML]{53B56D}57.8 & \cellcolor[HTML]{F4EFE4}-5.9 & \cellcolor[HTML]{E9F1FD}94.3 & \cellcolor[HTML]{7CC790}67.9 & \multicolumn{1}{c|}{\cellcolor[HTML]{F6E4B3}-26.4} & \cellcolor[HTML]{C5D9FB}84.8 & \cellcolor[HTML]{81C994}69.1 & \multicolumn{1}{c|}{\cellcolor[HTML]{F5EACD}-15.7} & \cellcolor[HTML]{C2D7FB}83.9 & \cellcolor[HTML]{4AB165}55.5 & \cellcolor[HTML]{F6E3AF}-28.4 \\ \midrule
{\tt Qwen2.5-7B} & \cellcolor[HTML]{FDFEFE}99.6 & \cellcolor[HTML]{B5DFC0}18.5 & \multicolumn{1}{c|}{\cellcolor[HTML]{FAC631}-81.2} & \cellcolor[HTML]{FCFDFE}99.3 & \cellcolor[HTML]{A4D8B2}22.5 & \multicolumn{1}{c|}{\cellcolor[HTML]{FAC83B}-76.7} & \cellcolor[HTML]{F0F5FE}96.1 & \cellcolor[HTML]{84CB97}30.3 & \cellcolor[HTML]{F9CE55}-65.8 & \cellcolor[HTML]{F8FAFE}98.2 & \cellcolor[HTML]{EBF7EE}5.2 & \multicolumn{1}{c|}{\cellcolor[HTML]{FBBF14}-93.1} & \cellcolor[HTML]{F8FAFE}98.2 & \cellcolor[HTML]{E0F2E5}7.8 & \multicolumn{1}{c|}{\cellcolor[HTML]{FBC11A}-90.4} & \cellcolor[HTML]{ABC9FA}77.9 & \cellcolor[HTML]{85CB98}30.2 & \cellcolor[HTML]{F7D880}-47.7 \\ \midrule
{\tt Qwen2.5-Math} & \cellcolor[HTML]{FEFEFE}99.8 & \cellcolor[HTML]{88CC9A}70.7 & \multicolumn{1}{c|}{\cellcolor[HTML]{F6E3AD}-29.1} & \cellcolor[HTML]{F9FBFE}98.5 & \cellcolor[HTML]{EFF8F1}96.1 & \multicolumn{1}{c|}{\cellcolor[HTML]{F4F1ED}-2.4} & \cellcolor[HTML]{F6F9FE}97.8 & \cellcolor[HTML]{F6FBF7}97.8 & \cellcolor[HTML]{F3F3F3}0.0 & \cellcolor[HTML]{95BAF8}72.0 & \cellcolor[HTML]{84CA97}69.9 & \multicolumn{1}{c|}{\cellcolor[HTML]{F4F1EE}-2.0} & \cellcolor[HTML]{76A6F7}{\color[HTML]{FFFFFF} \textbf{36.5}} & \cellcolor[HTML]{88CC9A}29.4 & \multicolumn{1}{c|}{\cellcolor[HTML]{F4EFE2}-7.1} & \cellcolor[HTML]{5B96F6}{\color[HTML]{FFFFFF} \textbf{43.4}} & \cellcolor[HTML]{4FB46A}43.4 & \cellcolor[HTML]{F3F3F3}0.0 \\ \midrule
{\tt GPT-3.5\_Turbo} & \cellcolor[HTML]{D5E4FC}89.0 & \cellcolor[HTML]{4AB266}{\color[HTML]{FFFFFF} \textbf{44.7}} & \multicolumn{1}{c|}{\cellcolor[HTML]{F7DA89}-44.3} & \cellcolor[HTML]{90B7F8}70.8 & \cellcolor[HTML]{D0EBD7}11.7 & \multicolumn{1}{c|}{\cellcolor[HTML]{F8D265}-59.1} & \cellcolor[HTML]{B1CCFA}79.5 & \cellcolor[HTML]{47B063}{\color[HTML]{FFFFFF} \textbf{45.5}} & \cellcolor[HTML]{F6E0A1}-34.0 & \cellcolor[HTML]{FDFEFE}99.7 & \cellcolor[HTML]{63BC7B}{\color[HTML]{FFFFFF} \textbf{61.8}} & \multicolumn{1}{c|}{\cellcolor[HTML]{F7DE98}-37.8} & \cellcolor[HTML]{F8FAFE}98.2 & \cellcolor[HTML]{74C389}65.9 & \multicolumn{1}{c|}{\cellcolor[HTML]{F6E1A5}-32.3} & \cellcolor[HTML]{87B1F8}68.3 & \cellcolor[HTML]{4FB46A}43.4 & \cellcolor[HTML]{F5E5B7}-24.9 \\ \midrule
{\tt o3-Mini} & \cellcolor[HTML]{FAFCFE}98.9 & \cellcolor[HTML]{F8FCF9}98.3 & \multicolumn{1}{c|}{\cellcolor[HTML]{F4F2F1}-0.6} & \cellcolor[HTML]{F9FBFE}98.5 & \cellcolor[HTML]{FCFEFD}99.5 & \multicolumn{1}{c|}{\cellcolor[HTML]{F3F3F1}+1.0} & \cellcolor[HTML]{ECF3FD}95.1 & \cellcolor[HTML]{FCFDFC}99.3 & \cellcolor[HTML]{F3F1E9}+4.2 & \cellcolor[HTML]{FCFDFE}99.3 & \cellcolor[HTML]{FEFEFE}99.9 & \multicolumn{1}{c|}{\cellcolor[HTML]{F3F3F2}+0.6} & \cellcolor[HTML]{FBFCFE}99.1 & \cellcolor[HTML]{F7FBF8}98.2 & \multicolumn{1}{c|}{\cellcolor[HTML]{F4F2F0}-0.9} & \cellcolor[HTML]{CCDEFC}86.7 & \cellcolor[HTML]{C9E8D1}86.8 & \cellcolor[HTML]{F3F3F3}+0.1 \\ \midrule
{\tt SocraticLM} & \cellcolor[HTML]{FCFDFE}99.3 & \cellcolor[HTML]{F1F9F3}96.8 & \multicolumn{1}{c|}{\cellcolor[HTML]{F4F1EC}-2.6} & \cellcolor[HTML]{FDFEFE}99.7 & \cellcolor[HTML]{FBFDFC}99.3 & \multicolumn{1}{c|}{\cellcolor[HTML]{F4F2F1}-0.5} & \cellcolor[HTML]{F3F7FE}97.0 & \cellcolor[HTML]{F8FCF9}98.4 & \cellcolor[HTML]{F3F3F0}+1.4 & \cellcolor[HTML]{679CF6}{\color[HTML]{FFFFFF} \textbf{59.8}} & \cellcolor[HTML]{6CC082}63.9 & \multicolumn{1}{c|}{\cellcolor[HTML]{F3F1EA}+4.1} & \cellcolor[HTML]{7FACF8}34.0 & \cellcolor[HTML]{61BC79}39.1 & \multicolumn{1}{c|}{\cellcolor[HTML]{F3F1E7}+5.0} & \cellcolor[HTML]{84B0F8}32.6 & \cellcolor[HTML]{37AA56}{\color[HTML]{FFFFFF} \textbf{49.3}} & \cellcolor[HTML]{F4EACC}+16.7 \\ \midrule
{\tt LearnLM-1.5-Pro} & \cellcolor[HTML]{FEFEFE}99.8 & \cellcolor[HTML]{9AD3A9}75.3 & \multicolumn{1}{c|}{\cellcolor[HTML]{F5E5B8}-24.6} & \cellcolor[HTML]{FDFEFE}99.7 & \cellcolor[HTML]{E4F3E8}93.6 & \multicolumn{1}{c|}{\cellcolor[HTML]{F4EFE4}-6.2} & \cellcolor[HTML]{FAFCFE}98.9 & \cellcolor[HTML]{F8FCF9}98.4 & \cellcolor[HTML]{F4F2F1}-0.5 & \cellcolor[HTML]{FEFEFE}99.9 & \cellcolor[HTML]{9CD5AB}24.5 & \multicolumn{1}{c|}{\cellcolor[HTML]{FAC93E}-75.4} & \cellcolor[HTML]{EBF2FD}94.9 & \cellcolor[HTML]{41AD5E}{\color[HTML]{FFFFFF} \textbf{53.3}} & \multicolumn{1}{c|}{\cellcolor[HTML]{F7DC8F}-41.6} & \cellcolor[HTML]{7BA9F7}65.1 & \cellcolor[HTML]{57B770}58.7 & \cellcolor[HTML]{F4EFE3}-6.4 \\ \midrule[1.5pt]
Avg. Deviation & 40.5 & 27.5 & \multicolumn{1}{c|}{-13.0} & 35.0 & 30.2 & \multicolumn{1}{c|}{-4.8} & 32.9 & 28.8 & -4.2 & 41.9 & 30.6 & \multicolumn{1}{c|}{-11.3} & 37.8 & 27.5 & \multicolumn{1}{c|}{-10.3} & 25.4 & 15.2 & -10.2 \\ \midrule
Random Choice & \multicolumn{2}{c}{6.1} & \multicolumn{1}{l|}{} & \multicolumn{2}{c}{1.4} & \multicolumn{1}{l|}{} & \multicolumn{2}{c}{6.7} & \multicolumn{1}{l|}{} & \multicolumn{2}{c}{4.12} & \multicolumn{1}{l|}{} & \multicolumn{2}{c}{4} & \multicolumn{1}{l|}{} & \multicolumn{2}{c}{0.9} & \multicolumn{1}{l}{} \\ \bottomrule[1.5pt]
\end{tabular}%
}
\caption{LLM percentile scores on grade-level questions from mathematics and reading without grade enforcement ($P_{U}$ -- shaded \highlighthex[0pt]{4285F4}{blue}), with grade enforcement ($P_{E}$ -- shaded \highlighthex[0pt]{34A853}{green}), and their difference ($\Delta$ = $P_{U}$ - $P_{E}$ -- shaded \highlighthex[0pt]{FBBC04}{yellow}). Darker hues for $P_{U}$ and $P_{E}$ denote closer alignment to the average score of 50 and larger absolute change in $\Delta$. {\bf Boldface} highlights the best model (i.e., closest to 50) in each setting. {\tt Avg. Deviation} records the mean absolute deviation from P=50 for corresponding prompt settings. The {\tt Random Choice} baseline reports the percentile scores attained with a randomized option chosen for each problem.}
\label{tab:main-results}
\end{table*}

\section{Results \& Discussion}
\label{sec:results}


\begin{itemize}
  \item address {\bf RQ1} by comparing model percentiles to the student mean (50\textsuperscript{th} percentile) across grades and subjects, and 
  \item address {\bf RQ2} by (i) quantifying the effect of grade enforcement on LLM performance ({\bf RQ2.1}) and (ii) evaluating whether these shifts mirror human student response patterns ({\bf RQ2.2}).
\end{itemize}

For further context, Table \ref{tab:unenforced-acc} presents the accuracy of each LLM under the unenforced condition.

\subsection{RQ1: Alignment under Unenforced Prompting}

We ask whether the unenforced problem solving prompt generates outputs that align with that of the average student in each grade (\highlighthex[0pt]{4285F4}{$P_{U}$}). 

\paragraph{Mathematics.}  
Most models, especially those scoring well on GSM8K, e.g., {\tt LLaMA3.1-70B}, {\tt Qwen2.5-Math}, {\tt o3-Mini}, and{\tt SocraticLM}, achieve high percentiles in every grade, overshooting all benchmarks and showing no alignment with any specific grade. This is also reflected in the high average deviation of 40.5, 35.0, and 32.9 percentile points, respectively, from the optimal P=50 mark. In contrast, smaller models with relatively poorer benchmark performance, such as {\tt LLaMA2-13B} and {\tt Mistral-7B}, exhibit lower percentiles and show better alignment between grades.

\paragraph{Reading.}  
Similar to mathematics, the models demonstrate high average percentile scores in reading for grades 4 and 8, proving unsuitable for faithful student mimicking. The models in our pool align better with grade 12, with relatively lower average \highlighthex[0pt]{4285F4}{$P_{U}$} values. Fine-tuned models (not tuned for grade-alignment), e.g., {\tt Qwen2.5-Math}, {\tt SocraticLM} -- {\tt Qwen2.5-Math} further tuned on pedagogical data, have a poorer overall performance, resulting in better alignment across grades.

Across grades and subjects, all models score well above the {\tt Random Choice} baseline. Without enforced instructions, LLMs rarely self-calibrate to grade difficulty. They overshoot when capacity is high and align only when under-powered or off-domain.

\subsection{RQ2.1: Effect of Grade-Level Prompts}



\paragraph{Gains:} We observe that grade-specific prompting can also increase model performance. For example, several settings with {\tt LLaMA2-13B/70B} for mathematics and all grade settings with {\tt SocraticLM} for reading result in higher $P_{E}$ than $P_{U}$.

\paragraph{Stable:} Some models, such as {\tt LLaMA3.1-70B} and o3-Mini for subjects and {\tt SocraticLM} for mathematics, show little to no change between their values of $P_{U}$ and $P_{E}$ values, despite their respective $P_{U}$ values having a high deviation from the target P=50.

\paragraph{Prompt Strength:} Among the three grade enforcement prompts, the most detailed \textsc{GradeEnforcedFullCoT} prompt (with explicit instruction to consider the probability that an average student of the target grade will get the given problem right) causes the largest changes (Figure \ref{fig:by-prompt-rq2.1}). This shows that grade-level cues can markedly increase or lower scores depending on the model and prompt strength, although a few models remain robust.

\subsection{RQ2.2: Alignment Under Enforced Prompts}

We investigate whether grade-specific prompts move the model performance closer to the average student (ideal $P=50$). We find that the results are spread across the following categories:

\paragraph{(1) Aligned $P_{U}$ and aligned $P_{E}$:}
Some models that are close to the 50th percentile without grade-specific prompting maintain good alignment after prompting (for example, {\tt LLaMA2-13B / 70B} for mathematics and {\tt SocraticLM} for reading). These models can act as “proxy students” out of the box for particular pairs of subjects' grades. 

\paragraph{(2) Misaligned $P_{U}$ and misaligned $P_{E}$:}
Other models' percentile scores can range far above or below the median despite grade-specific prompting (for example, $P_{U}$s and $P_{E}$s for {\tt o3-Mini} across subjects and grades stay far above 50). We did not observe any model that consistently scored below the median percentile. 

\paragraph{(3) Misaligned $P_{U}$ and aligned $P_{E}$:}
In some cases, prompting can help induce grade alignment ($P_{E}$) when unenforced alignment is poor ($P_{U}$). For example, {\tt Mistral-7B}'s percentile range on reading problems moves from $P_U \in [83.9-94.3]$ to $P_E{=}[55.5-69.1]$; {\tt GPT-3.5-Turbo} shows similar gains in most tasks. Such cases demonstrate the desired effect of grade-specific prompting.

\paragraph{(4) Aligned $P_{U}$ and misaligned $P_{E}$:}
In contrast, grade-specific prompting can cause models to overshoot. {\tt Qwen2.5-7B} in grade 4 reading drops from 98.2 to 5.2 ($\Delta{=}{-}$93.1), overshooting the target.

\begin{figure}[!ht]
  \centering

  \begin{subfigure}{\columnwidth}
    \centering
    \includegraphics[width=\linewidth]{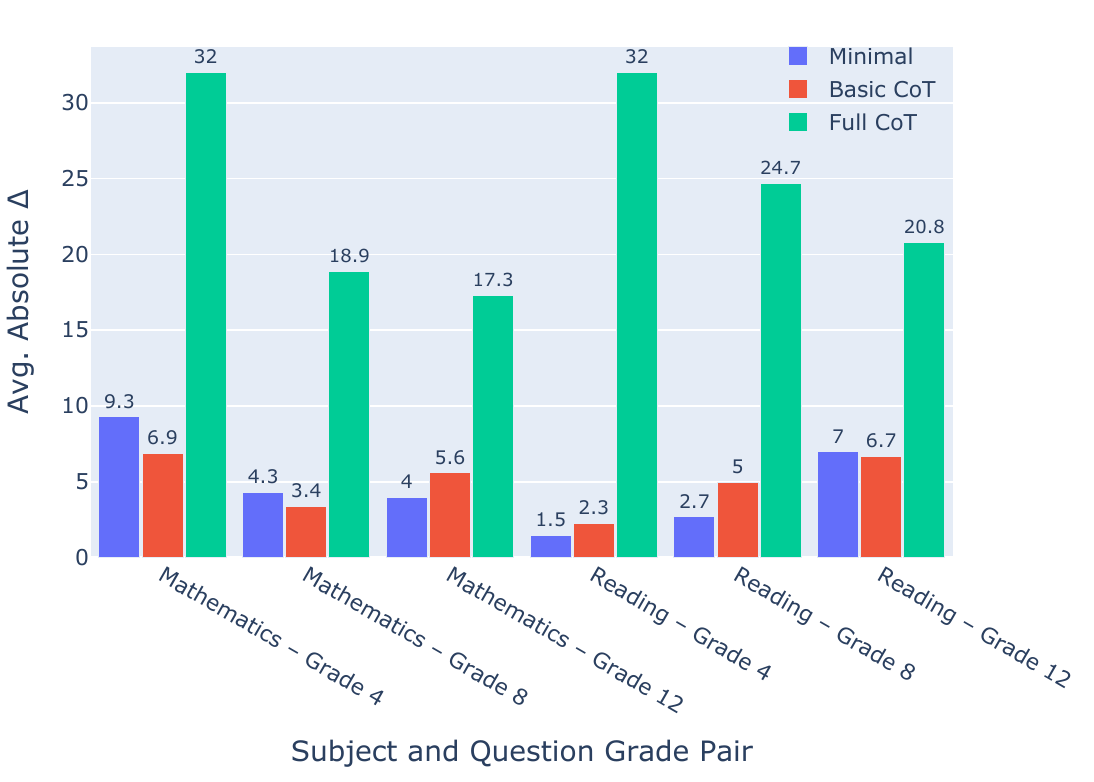}
    \caption{On $\Delta$}
    \label{fig:by-prompt-rq2.1}
  \end{subfigure}

  \begin{subfigure}{\columnwidth}
    \centering
    \includegraphics[width=\linewidth]{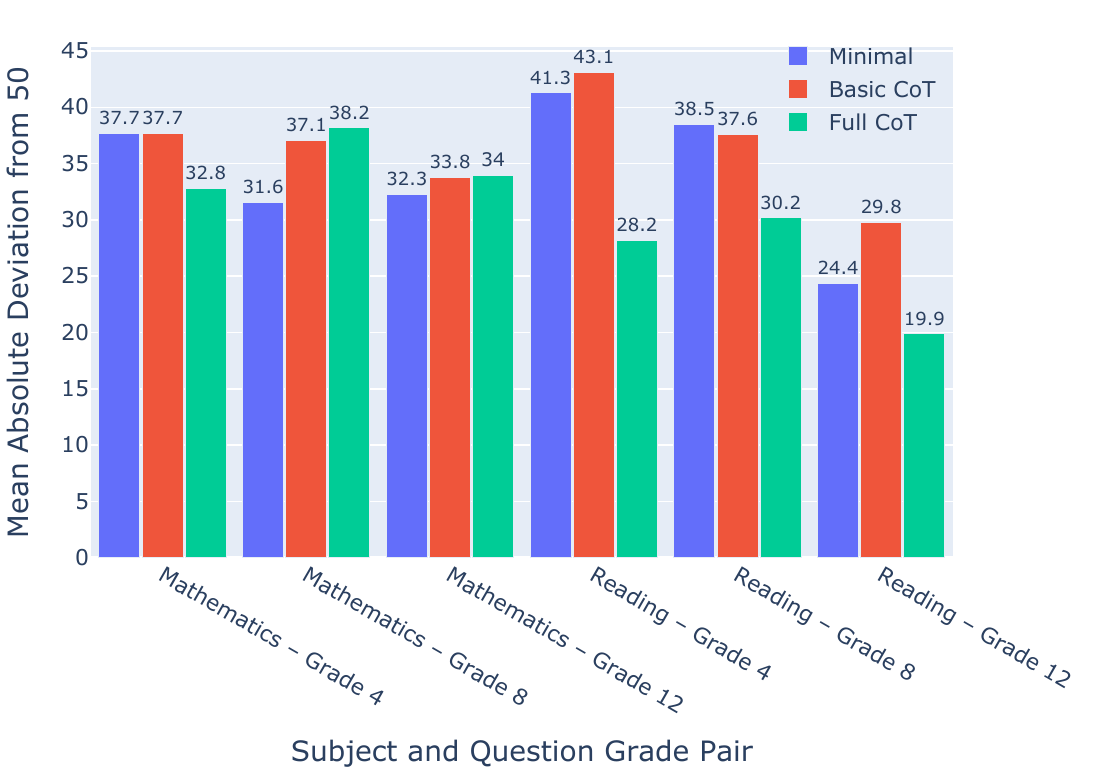}
    \caption{On grade-alignment. We report mean absolute deviation from an average grade-level student’s percentile score (i.e., 50). Thus, greater the deviation, poorer the average alignment.}
    \label{fig:by-prompt-rq2.2}
  \end{subfigure}

  \caption{Impact of grade-enforcing prompting}
  \label{fig:by-prompt-rq2}
\end{figure}

\paragraph{Prompt design matters.}
Figure~\ref{fig:by-prompt-rq2.2} shows that the \textsc{GradeEnforcedFullCoT} template changes scores the most. However, it is not always the most optimal prompt setting to achieve better grade alignment (lower percentile deviation from 50).

\paragraph{Fine-tuned models.}
Pedagogically tuned models ({\tt LearnLM-1.5-Pro}, and {\tt SocraticLM}) are not better aligned than general LLMs (such as {\tt Mistral-7B}), with or without prompts, indicating that faithful grade-level emulation probably needs explicit alignment objectives.

Thus, grade alignment is model-prompt specific; no single prompt works everywhere. Reliable grade-level emulation will require tailored prompting that does not ensure generalization to other grades or subjects.

\subsection{Guidelines for Selecting Viable LLM “Proxy Students”}

Our experiments confirm that no single model–prompt pair reliably matches average student performance in every grade and subject. Before an LLM can stand in for real students, e.g., to trial new test items or train a model for an ITS, it should pass the following baseline checks:

\begin{enumerate}
    \item \textbf{Grade alignment.} The model’s ability estimate ($\theta_n$) in a representative item set must fall within the normative band of the grade: core average $\pm1$ logit (percentile: 6.68 to 93.32), extended $\pm1.5$ (percentile: 2.28 to 97.72), outlier $\ge\pm2$ (percentile: 15.87 to 84.13) \citep{BondFox2015}. Models such as {\tt GPT-3.5-Turbo} stayed in the core range with appropriate prompts for most grades.

    \item \textbf{Developmental ordering.} Ability should rise monotonically with grade, mirroring trends in NAEP reading (217, 262, and 285 according to NAEP's own cumulative scoring scale for grades 4, 8, 12 respectively; \citealp{nces2022naep}). Several pairs violated this; e.g., {\tt Mistral-7B}'s $P_{U}$ was 38.3, 17.0, 40.5 for the same grades.

    \item \textbf{Prompt stability.} Grade-enforcing prompts can improve or harm performance. An unenforced prompt should be used if the model is already aligned; otherwise, one should verify that enforcement is equally accurate across all grades.
\end{enumerate}

\textit{These criteria are necessary, but not sufficient}. We believe that more accurate guidelines for faithful student mimicking will emerge with richer evaluation datasets.

\section{Conclusion}
\label{sec:conclusion}

In this paper, we investigate whether LLMs' regular problem-solving performance aligns with that of an average student of a given grade, and whether explicit prompting to act like the average student makes a difference and improves this alignment. We conduct a thorough analysis of 11 diverse models on mathematics and reading questions from K-12 grades 4, 8, and 12 sourced from the NAEP database. Our IRT-based analysis reveals that in the regular (unenforced) setting, stronger models score far better than the average students of any grade and weaker models may align well incidentally. Though explicit (grade-enforced) prompting causes a change in model performance, the alignment with the desired grade-level average varies substantially across model and prompt combinations, with no single model-prompt pair producing average performance across grades or subjects. We provide a set of necessary guidelines to select viable student-proxies for future work and highlight the need for dedicated model finetuning for faithful grade adherence.

\section*{Limitations}

While the results of our experiments lead to certain conclusions and provide us with novel insights, we acknowledge that these are necessarily limited in a number of ways.

\paragraph{Limited number of samples and subjects considered:} Getting access to publicly available student answering data is challenging. The NAEP \cite{nces2022naep} database offers a valuable resource in that regard. However, the database is not naturally designed to provide data for performing analysis over automated models at scale, therefore, the available subjects and the number of questions in the collected dataset are limited. 

\paragraph{Text-based questions only:} In this study, we have restricted our analysis to text-only questions, omitting questions that involve visual interpretation. We admit that this is not completely faithful to student assessments, as visual cues may also elicit key reasoning abilities. We plan  to expand our study to more modalities in the future.

\paragraph{MCQ format:} Evaluation of LLM responses is a key challenge, especially for free-form answering style. To mitigate this challenge, in this work, we focus on MCQ-type questions only. This also makes modeling the items within the IRT framework easier. As models vary in their response structure, we find that simple rule-based extraction is not reliable enough, and we have to use a follow-up prompt to extract the final option selected by the model. We plan to develop more robust evaluation strategies to allow for more varied question types in the future.

\paragraph{No data for cross-grade performance used:} A key point to note is that NAEP only reports the performance of students at a particular grade level on questions from the same grade. Though this is adequate for assessing student learning trends, for determining cross-grade viability of proxy students, we would require real students' performance on questions from different grades.

\paragraph{Use of prompting methods only:} We focus our study solely on prompt-based methods to enforce grade-level alignment, as this is one of the most accessible ways in which models are used in this context, as demonstrated by previous work. A more in-depth analysis is needed to assess whether in-context learning and finetuning strategies can also play a role in improving the quality of proxy-students, in addition to appropriately sized student demonstration data for tuning. We also highlight that prompt engineering (i.e., designing the most optimal prompts for the models) was outside the scope of this study, and the prompts that we used are inspired by previous work in this domain.

\paragraph{Experiments with specific models:} Last but not least, we acknowledge that our findings apply to a specific set of models considered in this study. We highlight that our choice was motivated by considerations around the diversity of the model pool.

\section*{Ethical Considerations}

This study relies exclusively on cumulative, de-identified statistics drawn from student response data supplied by the National Assessment of Educational Progress (NAEP). No record contains direct or indirect identifiers, and at no stage were individual-level student data accessed, stored, or analyzed. All analytic procedures conformed to the NAEP Data Confidentiality and Disclosure Policy as well as the privacy protections required under the Family Educational Rights and Privacy Act (FERPA). Consequently, the research poses no risk to the privacy or well-being of individual students.

\bibliography{custom}

\begin{thebibliography}{34}
\providecommand{\natexlab}[1]{#1}

\bibitem[{Baker(2001)}]{baker2001basics}
Frank~B Baker. 2001.
\newblock \emph{The basics of item response theory}.
\newblock ERIC.

\bibitem[{Belz et~al.(2023)Belz, Thomson, Reiter, and Mille}]{belz-etal-2023-non}
Anya Belz, Craig Thomson, Ehud Reiter, and Simon Mille. 2023.
\newblock \href {https://doi.org/10.18653/v1/2023.findings-acl.226} {Non-repeatable experiments and non-reproducible results: The reproducibility crisis in human evaluation in {NLP}}.
\newblock In \emph{Findings of the Association for Computational Linguistics: ACL 2023}, pages 3676--3687, Toronto, Canada. Association for Computational Linguistics.

\bibitem[{Benedetto et~al.(2024)Benedetto, Aradelli, Donvito, Lucchetti, Cappelli, and Buttery}]{benedetto-etal-2024-using}
Luca Benedetto, Giovanni Aradelli, Antonia Donvito, Alberto Lucchetti, Andrea Cappelli, and Paula Buttery. 2024.
\newblock \href {https://doi.org/10.18653/v1/2024.findings-emnlp.663} {Using {LLM}s to simulate students' responses to exam questions}.
\newblock In \emph{Findings of the Association for Computational Linguistics: EMNLP 2024}, pages 11351--11368, Miami, Florida, USA. Association for Computational Linguistics.

\bibitem[{Bond and Fox(2015)}]{BondFox2015}
Trevor~G. Bond and Christine~M. Fox. 2015.
\newblock \href {https://doi.org/10.4324/9781315814698} {\emph{Applying the Rasch Model: Fundamental Measurement in the Human Sciences}}, 3 edition.
\newblock Routledge, New York.

\bibitem[{Chang et~al.(2024)Chang, Wang, Wang, Wu, Yang, Zhu, Chen, Yi, Wang, Wang et~al.}]{chang2024survey}
Yupeng Chang, Xu~Wang, Jindong Wang, Yuan Wu, Linyi Yang, Kaijie Zhu, Hao Chen, Xiaoyuan Yi, Cunxiang Wang, Yidong Wang, and 1 others. 2024.
\newblock A survey on evaluation of large language models.
\newblock \emph{ACM transactions on intelligent systems and technology}, 15(3):1--45.

\bibitem[{Cobbe et~al.(2021)Cobbe, Kosaraju, Bavarian, Chen, Jun, Kaiser, Plappert, Tworek, Hilton, Nakano et~al.}]{cobbe2021training}
Karl Cobbe, Vineet Kosaraju, Mohammad Bavarian, Mark Chen, Heewoo Jun, Lukasz Kaiser, Matthias Plappert, Jerry Tworek, Jacob Hilton, Reiichiro Nakano, and 1 others. 2021.
\newblock Training verifiers to solve math word problems.
\newblock \emph{arXiv preprint arXiv:2110.14168}.

\bibitem[{Embretson and Reise(2000)}]{EmbretsonReise2000}
Susan~E. Embretson and Steven~P. Reise. 2000.
\newblock \href {https://doi.org/10.4324/9781410605269} {\emph{Item Response Theory for Psychologists}}.
\newblock Multivariate Applications Series. Lawrence Erlbaum Associates, Mahwah, NJ.

\bibitem[{Grohs et~al.(2024)Grohs, Abb, Elsayed, and Rehse}]{Rehse2024}
Michael Grohs, Luka Abb, Nourhan Elsayed, and Jana{-}Rebecca Rehse. 2024.
\newblock Large language models can accomplish business process management tasks.
\newblock In \emph{Proceedings of the International Conference on Business Process Management}.
\newblock Extended version available as arXiv:2307.09923.

\bibitem[{He-Yueya et~al.(2024)He-Yueya, Ma, Gandhi, Domingue, Brunskill, and Goodman}]{heyueya2024psychometricalignmentcapturinghuman}
Joy He-Yueya, Wanjing~Anya Ma, Kanishk Gandhi, Benjamin~W. Domingue, Emma Brunskill, and Noah~D. Goodman. 2024.
\newblock \href {https://arxiv.org/abs/2407.15645} {Psychometric alignment: Capturing human knowledge distributions via language models}.
\newblock \emph{Preprint}, arXiv:2407.15645.

\bibitem[{Hendrycks et~al.(2020)Hendrycks, Burns, Basart, Zou, Mazeika, Song, and Steinhardt}]{hendrycks2020measuring}
Dan Hendrycks, Collin Burns, Steven Basart, Andy Zou, Mantas Mazeika, Dawn Song, and Jacob Steinhardt. 2020.
\newblock Measuring massive multitask language understanding.
\newblock \emph{arXiv preprint arXiv:2009.03300}.

\bibitem[{Jiang et~al.(2023)Jiang, Sablayrolles, Mensch, Bamford et~al.}]{mistral}
Albert~Q. Jiang, Alexandre Sablayrolles, Arthur Mensch, Chris Bamford, and 1 others. 2023.
\newblock Mistral 7b.
\newblock \emph{arXiv preprint arXiv:2310.06825}.

\bibitem[{Kim et~al.(2024)Kim, Yang, and Jung}]{KimJung2024}
Junseok Kim, Nakyeong Yang, and Kyomin Jung. 2024.
\newblock \href {https://doi.org/10.48550/arXiv.2408.08631} {Persona is a double-edged sword: Mitigating the negative impact of role-playing prompts in zero-shot reasoning tasks}.
\newblock \emph{arXiv preprint}.

\bibitem[{Liu et~al.(2024)Liu, Huang, Xiao, Sha, Wu, Liu, Wang, and Chen}]{socraticlm_2024}
Jiayu Liu, Zhenya Huang, Tong Xiao, Jing Sha, Jinze Wu, Qi~Liu, Shijin Wang, and Enhong Chen. 2024.
\newblock Socraticlm: Exploring socratic personalized teaching with large language models.
\newblock In \emph{Advances in Neural Information Processing Systems (NeurIPS) 2024}.

\bibitem[{Liu et~al.(2025)Liu, Bhandari, and Pardos}]{Liu2025}
Yunting Liu, Shreya Bhandari, and Zachary~A. Pardos. 2025.
\newblock \href {https://doi.org/10.1111/bjet.13570} {Leveraging llm respondents for item evaluation: A psychometric analysis}.
\newblock \emph{British Journal of Educational Technology}, 56:1028--1052.

\bibitem[{Macina et~al.(2023)Macina, Daheim, Chowdhury, Sinha, Kapur, Gurevych, and Sachan}]{macina-etal-2023-mathdial}
Jakub Macina, Nico Daheim, Sankalan Chowdhury, Tanmay Sinha, Manu Kapur, Iryna Gurevych, and Mrinmaya Sachan. 2023.
\newblock \href {https://doi.org/10.18653/v1/2023.findings-emnlp.372} {{M}ath{D}ial: A dialogue tutoring dataset with rich pedagogical properties grounded in math reasoning problems}.
\newblock In \emph{Findings of the Association for Computational Linguistics: EMNLP 2023}, pages 5602--5621, Singapore. Association for Computational Linguistics.

\bibitem[{Matsuda et~al.(2023)Matsuda, Lv, and Zheng}]{Matsuda2023}
Noboru Matsuda, Dan Lv, and Guoliang Zheng. 2023.
\newblock \href {https://doi.org/10.1007/s40593-022-00306-1} {Teaching how to teach promotes learning by teaching}.
\newblock \emph{International Journal of Artificial Intelligence in Education}, 33(3):720--751.

\bibitem[{Modi and the LearnLM~Team(2024)}]{learnlm_15_pro}
Abhinit Modi and the LearnLM~Team. 2024.
\newblock Learnlm: Improving gemini for learning.
\newblock \emph{arXiv preprint arXiv:2412.16429}.

\bibitem[{Mollick et~al.(2024)Mollick, Mollick, Bach, Ciccarelli, Przystanski, and Ravipinto}]{WhartonLLMClass2024}
Ethan~R. Mollick, Lilach Mollick, Natalie Bach, L.~J. Ciccarelli, Ben Przystanski, and Daniel Ravipinto. 2024.
\newblock \href {https://doi.org/10.48550/arXiv.2407.12796} {Ai agents and education: Simulated practice at scale}.
\newblock \emph{arXiv preprint}.

\bibitem[{{National Center for Education Statistics}(2022)}]{nces2022naep}
{National Center for Education Statistics}. 2022.
\newblock The nation's report card: 2022 naep reading and mathematics assessments.
\newblock \url{https://nces.ed.gov/nationsreportcard/}.
\newblock Accessed: 2025-04-20.

\bibitem[{OpenAI(2023)}]{openai2023gpt3_5_turbo}
OpenAI. 2023.
\newblock {GPT-3.5-Turbo} [large language model].
\newblock \url{https://platform.openai.com/docs/models/gpt-3-5-turbo}.
\newblock Accessed: 2025-04-25.

\bibitem[{OpenAI(2025)}]{openai2025o3_mini}
OpenAI. 2025.
\newblock {OpenAI o3-mini} [large language model].
\newblock \url{https://platform.openai.com/docs/models/o3-mini}.
\newblock Accessed: 2025-04-25.

\bibitem[{Rasch(1960)}]{rasch1960probabilistic}
Georg Rasch. 1960.
\newblock \emph{Probabilistic Models for Some Intelligence and Attainment Tests}.
\newblock Danish Institute for Educational Research, Copenhagen.
\newblock Reprinted by University of Chicago Press (1980) and MESA Press (1992).

\bibitem[{Smith et~al.(2024)Smith, Gupta, and MacLellan}]{MacLellan2024}
Glen Smith, Adit Gupta, and Christopher~J. MacLellan. 2024.
\newblock \href {https://arxiv.org/abs/2404.07883} {Apprentice tutor builder: A platform for users to create and personalize intelligent tutors}.
\newblock \emph{arXiv preprint}.

\bibitem[{Sonkar et~al.(2024)Sonkar, Chen, Liu, Baraniuk, and Sachan}]{sonkar2024llm}
Shashank Sonkar, Xinghe Chen, Naiming Liu, Richard~G Baraniuk, and Mrinmaya Sachan. 2024.
\newblock Llm-based cognitive models of students with misconceptions.
\newblock \emph{arXiv preprint arXiv:2410.12294}.

\bibitem[{Touvron et~al.(2023)Touvron, Martin, Stone, Albert, Almahairi et~al.}]{LLaMA2}
Hugo Touvron, Louis Martin, Kevin Stone, Peter Albert, Amjad Almahairi, and 1 others. 2023.
\newblock Llama 2: Open foundation and fine-tuned chat models.
\newblock \emph{arXiv preprint arXiv:2307.09288}.

\bibitem[{Tu et~al.(2024)Tu, Fan, Tian, Shen, Shang, Gao, and Yan}]{Tu2024}
Quan Tu, Shilong Fan, Zihang Tian, Tianhao Shen, Shuo Shang, Xin Gao, and Rui Yan. 2024.
\newblock \href {https://doi.org/10.18653/v1/2024.acl-long.638} {{C}haracter{E}val: A {C}hinese benchmark for role-playing conversational agent evaluation}.
\newblock In \emph{Proceedings of the 62nd Annual Meeting of the Association for Computational Linguistics (Volume 1: Long Papers)}, pages 11836--11850, Bangkok, Thailand. Association for Computational Linguistics.

\bibitem[{{UNESCO}(2023)}]{unesco2023tech}
{UNESCO}. 2023.
\newblock Global education monitoring report 2023: Technology in education.
\newblock \url{https://unesdoc.unesco.org/ark:/48223/pf0000385723}.

\bibitem[{{U.S. Department of Education}(2023)}]{ed2023ai}
{U.S. Department of Education}. 2023.
\newblock Artificial intelligence and the future of teaching and learning: Insights and recommendations.
\newblock \url{https://www.ed.gov/sites/ed/files/documents/ai-report/ai-report.pdf}.

\bibitem[{Wang et~al.(2024)Wang, Xu, Li, Zhang, Liang, Tang, Yu, and Wen}]{wang2024large}
Shen Wang, Tianlong Xu, Hang Li, Chaoli Zhang, Joleen Liang, Jiliang Tang, Philip~S Yu, and Qingsong Wen. 2024.
\newblock Large language models for education: A survey and outlook.
\newblock \emph{arXiv preprint arXiv:2403.18105}.

\bibitem[{Williams(2003)}]{williams2003skills}
Joel Williams. 2003.
\newblock \emph{The Skills for Life survey: A national needs and impact survey of literacy, numeracy and ICT skills}.
\newblock 490. The Stationery Office.

\bibitem[{Woolf et~al.(2013)Woolf, Arroyo et~al.}]{woolf2013developing}
Beverly Woolf, Ivon Arroyo, and 1 others. 2013.
\newblock Intelligent tutoring systems by and for the developing world: A review of trends and opportunities.
\newblock \emph{International Journal of Artificial Intelligence in Education}, 24(3):331--367.

\bibitem[{Yang et~al.(2024)Yang, Yang, Zhang, Hui et~al.}]{qwen25}
An~Yang, Baosong Yang, Beichen Zhang, Binyuan Hui, and 1 others. 2024.
\newblock Qwen2.5 technical report.
\newblock \emph{arXiv preprint arXiv:2412.15115}.

\bibitem[{Zelikman et~al.(2023)Zelikman, Ma, Tran, Yang, Yeatman, and Haber}]{zelikman-etal-2023-generating}
Eric Zelikman, Wanjing Ma, Jasmine Tran, Diyi Yang, Jason Yeatman, and Nick Haber. 2023.
\newblock \href {https://doi.org/10.18653/v1/2023.emnlp-main.135} {Generating and evaluating tests for k-12 students with language model simulations: A case study on sentence reading efficiency}.
\newblock In \emph{Proceedings of the 2023 Conference on Empirical Methods in Natural Language Processing}, pages 2190--2205, Singapore. Association for Computational Linguistics.

\bibitem[{Zhao et~al.(2024)Zhao, Chen, Yang, Liu, Deng, Cai, Wang, Yin, and Du}]{zhao2024explainability}
Haiyan Zhao, Hanjie Chen, Fan Yang, Ninghao Liu, Huiqi Deng, Hengyi Cai, Shuaiqiang Wang, Dawei Yin, and Mengnan Du. 2024.
\newblock Explainability for large language models: A survey.
\newblock \emph{ACM Transactions on Intelligent Systems and Technology}, 15(2):1--38.

\end{thebibliography}

\appendix

\section{Dataset Details}
\label{sec:app-dataset-details}
\begin{figure}[!ht]
    \centering
    \includegraphics[width=0.8\linewidth]{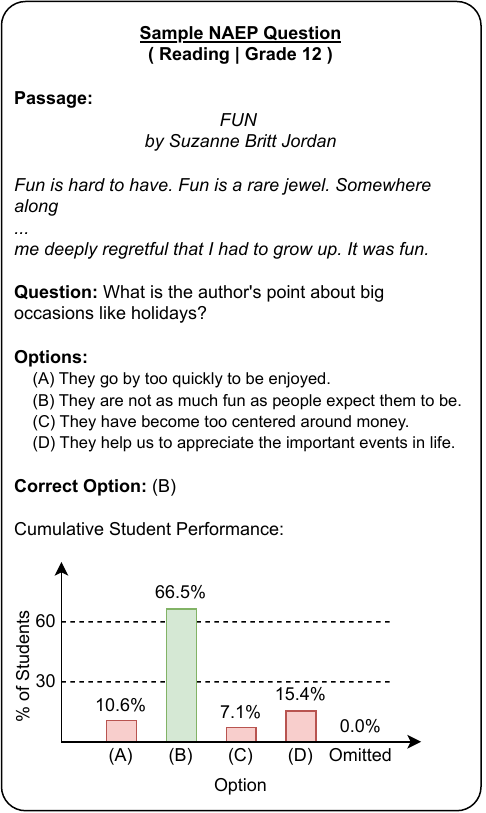}
    \caption{Sample NAEP question from grade 12 reading.}
    \label{fig:naep-example-g12-reading}
\end{figure}

Table \ref{tab:data-stats} presents the distribution of the questions extracted from NAEP by subject and grade levels. Figure \ref{fig:naep-example-g12-reading} shows an example NAEP question from grade 12 reading. We use data that is publicly available on the NAEP website. 

\begin{table}[!ht]
\centering
\resizebox{\columnwidth}{!}{%
\begin{tabular}{@{}lccc@{}}
\toprule
\textbf{Subject} & \textbf{Grade} & \textbf{Number of Questions} & \textbf{Percentage Share} \\ \midrule
\multirow{4}{*}{Mathematics} & 4 & 82 & 32.93 \\ \cmidrule(l){2-4} 
 & 8 & 106 & 42.57 \\ \cmidrule(l){2-4} 
 & 12 & 61 & 24.50 \\ \cmidrule(l){2-4} 
 & Total & 249 & 100.00 \\ \midrule
\multirow{4}{*}{Reading} & 4 & 101 & 42.08 \\ \cmidrule(l){2-4} 
 & 8 & 72 & 30.00 \\ \cmidrule(l){2-4} 
 & 12 & 67 & 27.92 \\ \cmidrule(l){2-4} 
 & Total & 240 & 100.00 \\ \bottomrule
\end{tabular}%
}
\caption{Dataset statistics: Number of Questions across Subjects and Grade-Levels}
\label{tab:data-stats}
\end{table}

\section{Prompt Templates}
\label{sec:app-prompts}

\begin{figure}[!ht]
    \centering
    \includegraphics[width=0.7\columnwidth]{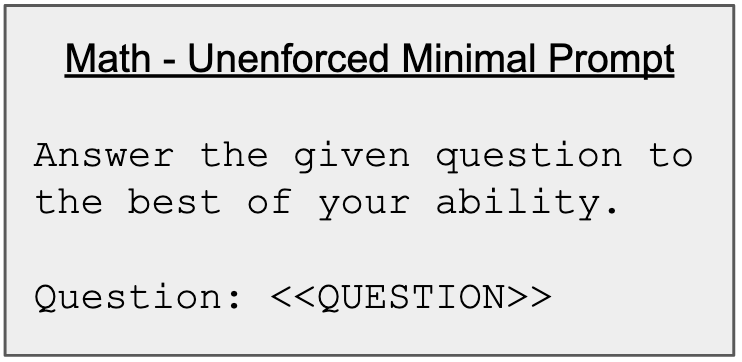}
    \caption{\textsc{Unenforced} Prompt Template for Mathematics}
    \label{fig:math-minimal-unenforced}
\end{figure}

\begin{figure}[!ht]
    \centering
    \includegraphics[width=0.7\columnwidth]{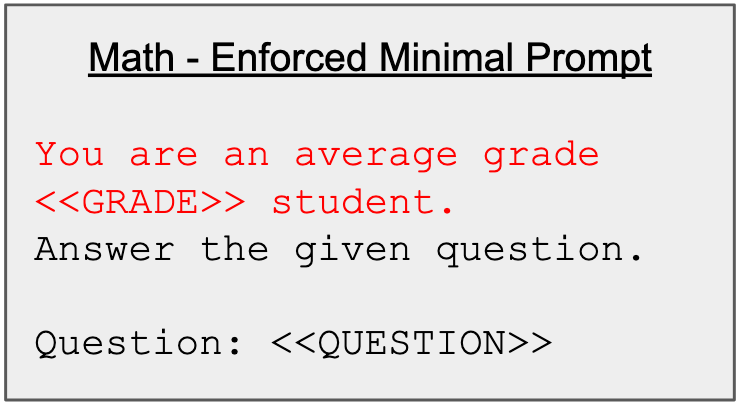}
    \caption{\textsc{GradeEnforcedMinimal} Prompt Template for Mathematics}
    \label{fig:math-minimal-enforced}
\end{figure}

\begin{figure}[!ht]
    \centering
    \includegraphics[width=0.7\columnwidth]{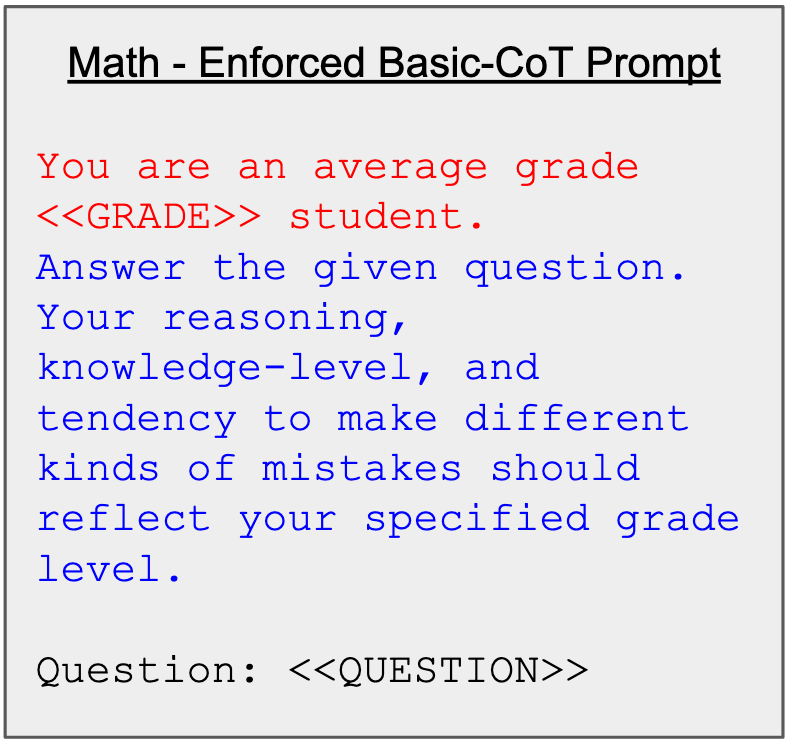}
    \caption{\textsc{GradeEnforcedBasicCoT} Prompt Template for Mathematics}
    \label{fig:math-basic-cot-enforced}
\end{figure}

\begin{figure}[!ht]
    \centering
    \includegraphics[width=0.7\columnwidth]{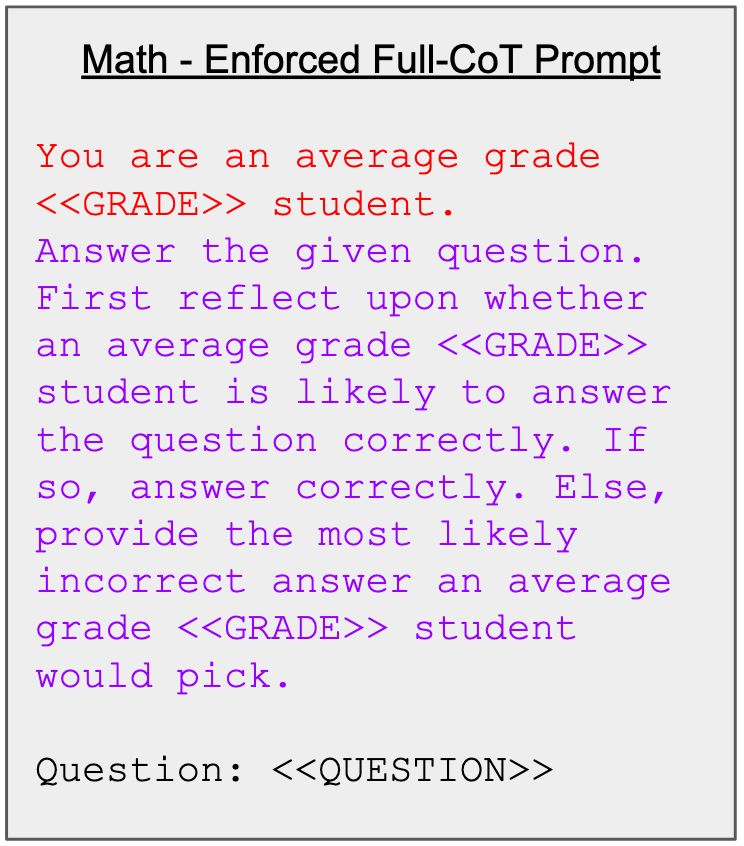}
    \caption{\textsc{GradeEnforcedFullCoT} Prompt Template for Mathematics}
    \label{fig:math-full-cot-enforced}
\end{figure}

\begin{figure}[!ht]
    \centering
    \includegraphics[width=0.7\columnwidth]{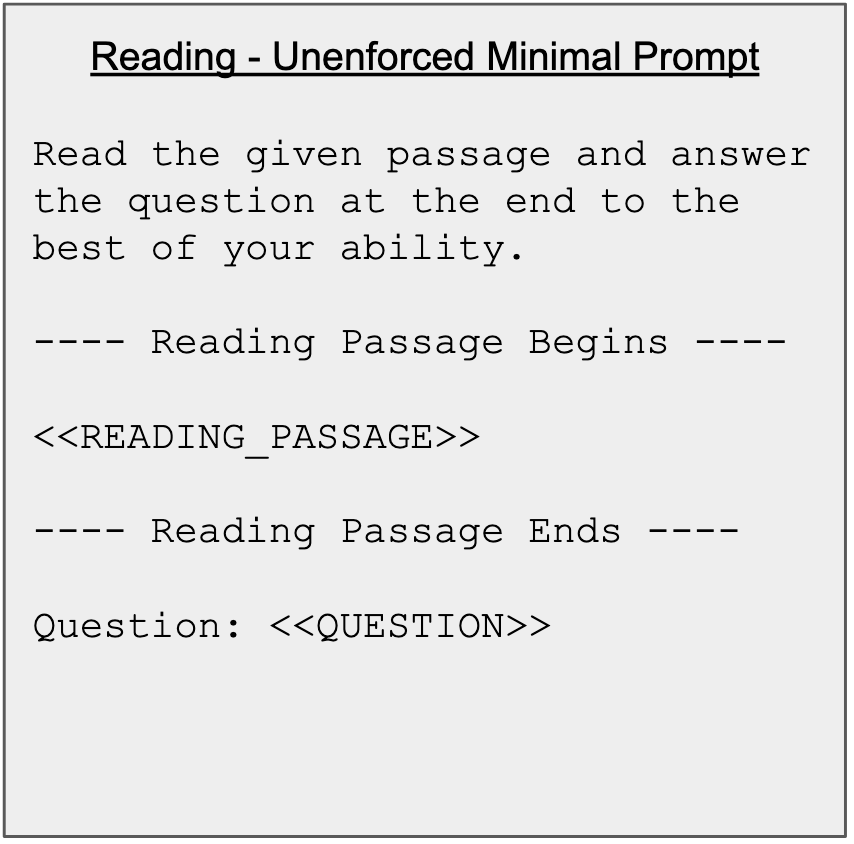}
    \caption{\textsc{Unenforced} Prompt Template for Reading}
    \label{fig:reading-minimal-unenforced}
\end{figure}

\begin{figure}[!ht]
    \centering
    \includegraphics[width=0.7\columnwidth]{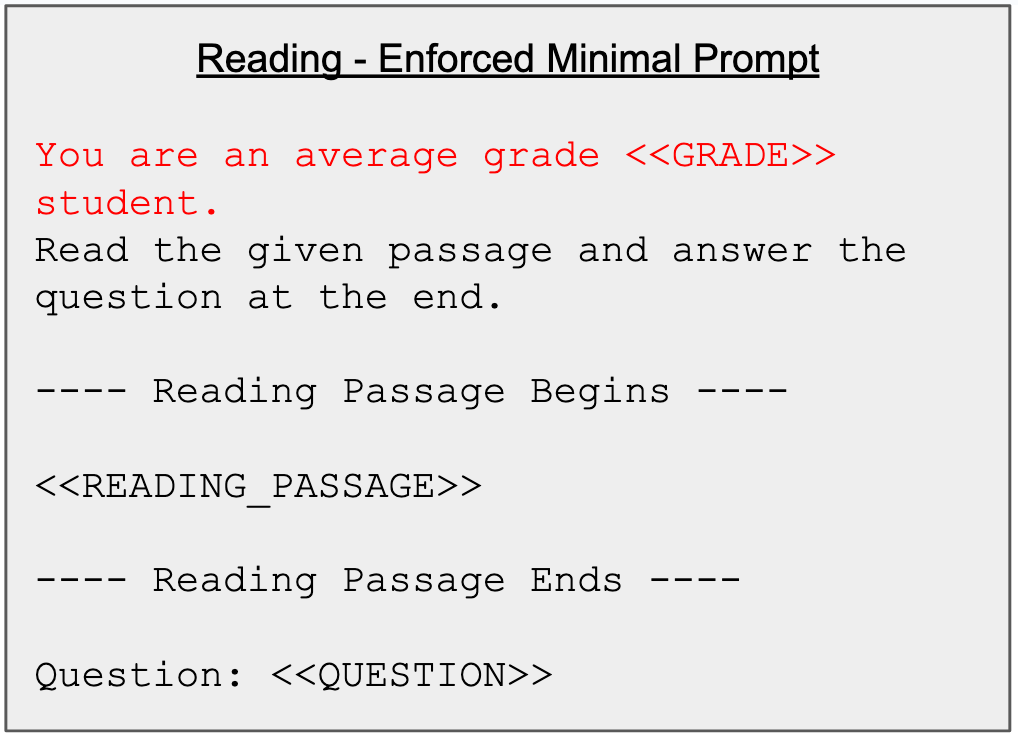}
    \caption{\textsc{GradeEnforcedMinimal} Prompt Template for Reading}
    \label{fig:reading-minimal-enforced}
\end{figure}

\begin{figure}[!ht]
    \centering
    \includegraphics[width=0.7\columnwidth]{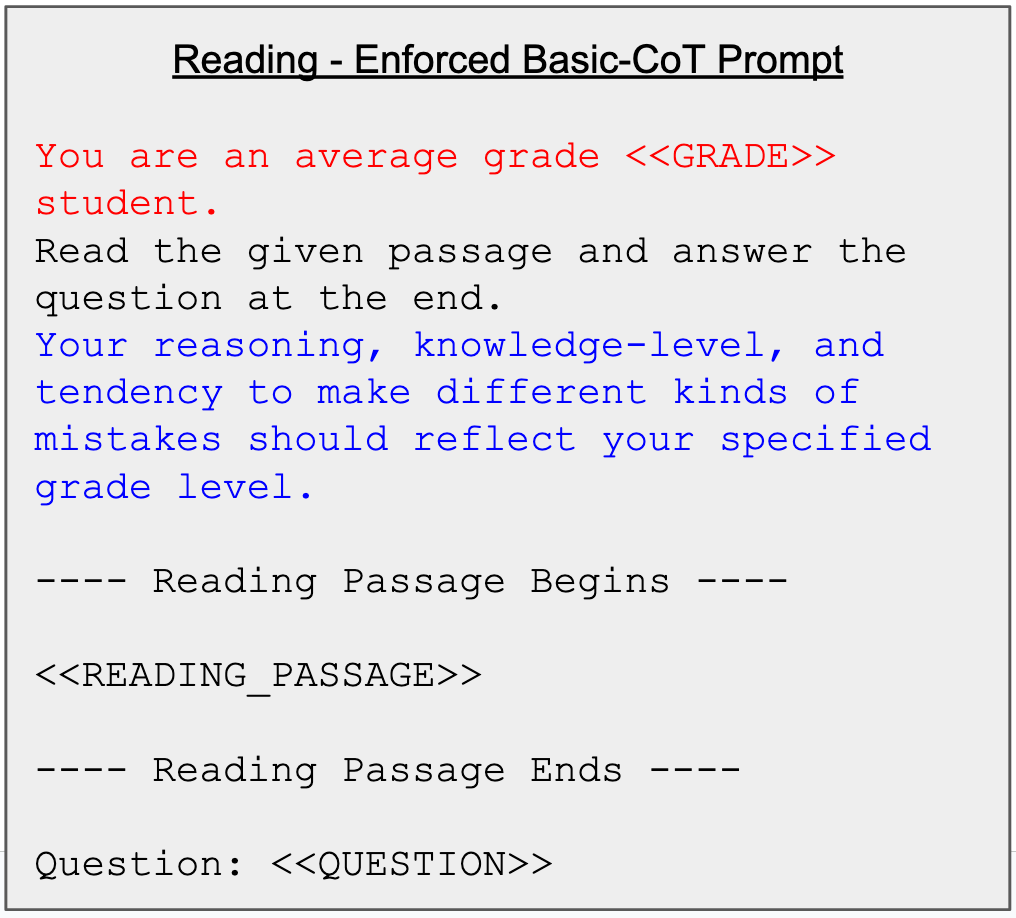}
    \caption{\textsc{GradeEnforcedBasicCoT} Prompt Template for Reading}
    \label{fig:reading-basic-cot-enforced}
\end{figure}

\begin{figure}[!ht]
    \centering
    \includegraphics[width=0.7\columnwidth]{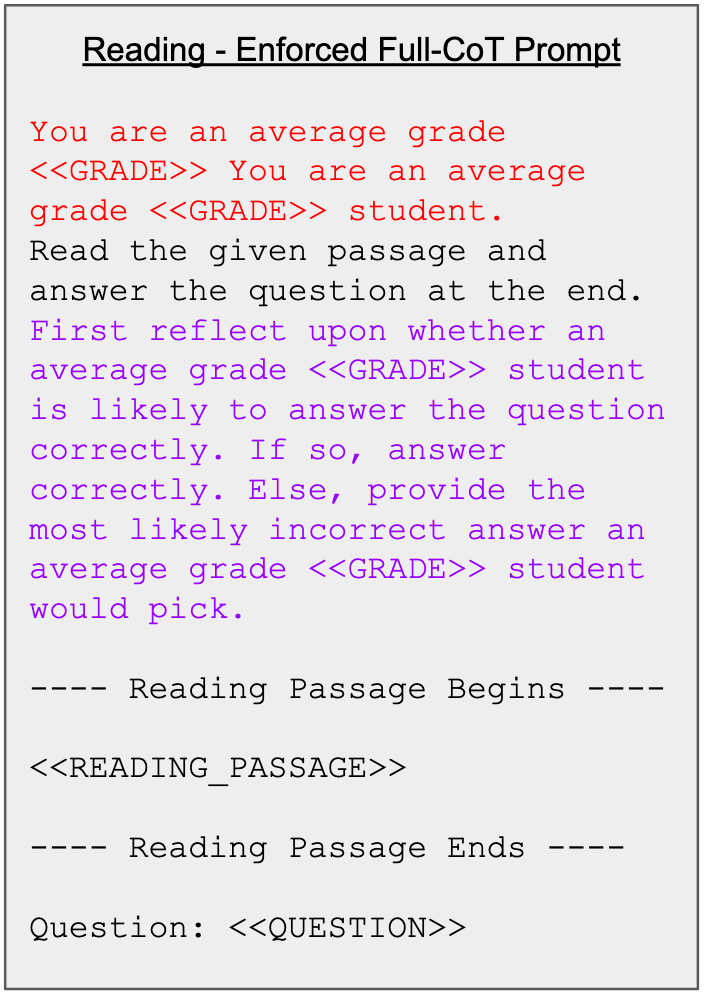}
    \caption{\textsc{GradeEnforcedFullCoT} Prompt Template for Reading}
    \label{fig:reading-full-cot-enforced}
\end{figure}

\begin{figure}[!ht]
    \centering
    \includegraphics[width=0.7\columnwidth]{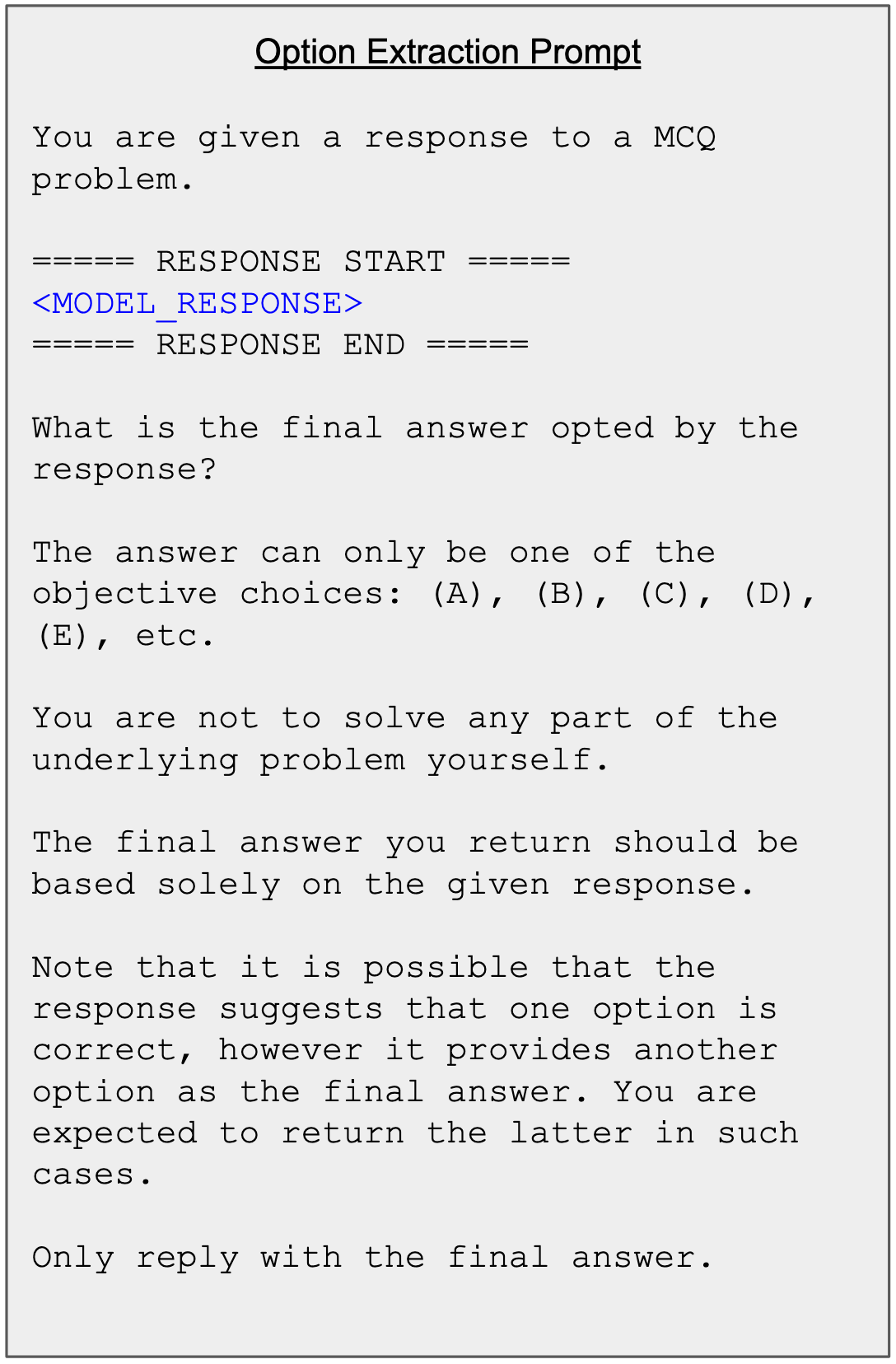}
    \caption{Option Extraction Prompt}
    \label{fig:option-extraction}
\end{figure}

Figures \ref{fig:math-minimal-unenforced} to \ref{fig:reading-full-cot-enforced} show the different solution generation prompts for mathematics and reading. Figure \ref{fig:option-extraction} shows the prompt used to extract the final option from the generated solution.

\section{Querying Setup}
All models were queried with the following hyperparameters: {\tt temperature}=0, {\tt top\_p}=0.95, and {\tt max\_tokens}=2048. {\tt LLaMA3.1} models were queried using the Google Cloud (Vertex) API, {\tt o3-Mini} was queried using the OpenAI API, and {\tt LearnLM-1.5-Pro} was queried using Google's \href{https://aistudio.google.com/prompts/new_chat}{AI Studio API}. All other models were imported from \href{https://huggingface.co/}{HunggingFace} and queried locally using \href{https://docs.vllm.ai/en/latest/}{vLLM} on a single NVIDIA A100 GPU. Each round of querying took less than one hour.

\section{Model Ability Estimation Algorithm}
\label{sec:app-theta-algo}

Algorithm \ref{alg:theta-estimation} captures the steps required to fit the Rasch model as described in §\ref{sec:methodology}.

\section{Analyses Details}
Table \ref{tab:unenforced-acc} lists LLMs' accuracy on mathematics and reading problems from different grade levels with the unenforced prompt setting. Table \ref{tab:best-prompt} records the best prompt out of the four possible settings, depending on the closeness of the corresponding percentile values to 50.

\begin{table}[!ht]
\centering
\resizebox{\columnwidth}{!}{%
\begin{tabular}{@{}lcccccc@{}}
\toprule
\multirow{2}{*}{\textbf{LLM}} & \multicolumn{3}{c}{\textbf{Mathematics}} & \multicolumn{3}{c}{\textbf{Reading}} \\ \cmidrule(l){2-7} 
 & \textbf{4} & \textbf{8} & \textbf{12} & \textbf{4} & \textbf{8} & \textbf{12} \\ \midrule
{\tt LLaMA2-13B} & 78.05 & 43.40 & 41.67 & 85.15 & 80.56 & 77.61 \\ \midrule
{\tt LLaMA2-70B} & 65.85 & 59.43 & 45.00 & 96.04 & 91.67 & 82.09 \\ \midrule
{\tt LLaMA3.1-8B} & 87.80 & 77.36 & 63.33 & 96.04 & 93.06 & 85.07 \\ \midrule
{\tt LLaMA3.1-70B} & 93.90 & 91.51 & 80.00 & 98.02 & 93.06 & 83.58 \\ \midrule
{\tt Mistral-7B} & 65.85 & 54.72 & 53.33 & 89.11 & 86.11 & 83.58 \\ \midrule
{\tt Qwen2.5-7B} & 93.90 & 92.45 & 80.00 & 93.07 & 94.44 & 80.60 \\ \midrule
{\tt Qwen2.5-Math} & 95.12 & 90.57 & 83.33 & 76.24 & 63.89 & 64.18 \\ \midrule
{\tt GPT-3.5\_Turbo} & 80.49 & 68.87 & 63.33 & 96.04 & 94.44 & 76.12 \\ \midrule
{\tt o3-Mini} & 91.46 & 90.57 & 78.33 & 95.05 & 95.83 & 85.07 \\ \midrule
{\tt SocraticLM} & 92.68 & 94.34 & 81.67 & 70.30 & 62.50 & 58.21 \\ \midrule
{\tt LearnLM-1.5-Pro} & 95.12 & 94.34 & 86.67 & 97.03 & 91.67 & 74.63 \\ \midrule
Average & 78.69 & 72.13 & 64.06 & 83.01 & 79.60 & 71.90 \\ \bottomrule
\end{tabular}%
}
\caption{LLM accuracy scores (i.e., accuracy in solving the tasks) for different grade levels in mathematics and reading under the unenforced prompt setting.}
\label{tab:unenforced-acc}
\end{table}

\begin{table}[!ht]
\centering
\resizebox{\columnwidth}{!}{%
\begin{tabular}{@{}lll@{}}
\toprule
\textbf{LLM} & \textbf{Best Prompt for Mathematics} & \textbf{Best Prompt for Reading} \\ \midrule
{\tt LLaMA2-13B} & \textsc{GradeEnforcedMinimal} & \textsc{GradeEnforcedMinimal} \\ \midrule
{\tt LLaMA2-70B} & \textsc{GradeEnforcedBasicCoT} & \textsc{GradeEnforcedFullCoT} \\ \midrule
{\tt LLaMA3.1-8B} & \textsc{GradeEnforcedMinimal} & \textsc{GradeEnforcedFullCoT} \\ \midrule
{\tt LLaMA3.1-70B} & \textsc{GradeEnforcedBasicCoT} & \textsc{GradeEnforcedMinimal} \\ \midrule
{\tt Mistral-7B} & \textsc{GradeEnforcedMinimal} & \textsc{GradeEnforcedFullCoT} \\ \midrule
{\tt Qwen2.5-7B} & \textsc{GradeEnforcedFullCoT} & \textsc{GradeEnforcedFullCoT} \\ \midrule
{\tt Qwen2.5-Math} & \textsc{GradeEnforcedFullCoT} & \textsc{GradeEnforcedMinimal} \\ \midrule
{\tt GPT-3.5-Turbo} & \textsc{GradeEnforcedFullCoT} & \textsc{GradeEnforcedFullCoT} \\ \midrule
{\tt o3-Mini} & \textsc{GradeEnforcedFullCoT} & \textsc{GradeEnforcedFullCoT} \\ \midrule
{\tt SocraticLM} & \textsc{GradeEnforcedFullCoT} & \textsc{GradeEnforcedFullCoT} \\ \midrule
{\tt LearnLM-1.5-Pro} & \textsc{GradeEnforcedFullCoT} & \textsc{GradeEnforcedFullCoT} \\ \bottomrule
\end{tabular}%
}
\caption{Best prompts for LLM in each subject. We pick the best prompt based on the closest average percentile rank to 50, i.e., the desired average performance.}
\label{tab:best-prompt}
\end{table}

\begin{algorithm*}
\caption{\small Estimating LLM Ability and Percentile Rank Using the Rasch Model}
\label{alg:theta-estimation}
\small
\begin{algorithmic}[1]
    \Require $p = \{p_j\}_{j=1}^I$ \Comment{Proportion of correct responses for item $j$ across students}
    \Require $s = \{s_{ij}\}_{i=1,j=1}^{M,I}$ \Comment{Binary correctness matrix: LLM $i$'s response to item $j$}
    \Ensure $\theta = \{\theta_i\}_{i=1}^M$ \Comment{Estimated ability (logit scale) for each LLM}
    \Ensure $\pi = \{\pi_i\}_{i=1}^M$ \Comment{Percentile rank of each LLM }

    \Statex \textbf{Step 1: Estimate item difficulties using student response proportions}

       \For{$j = 1$ to $I$}
        \State $b_j \gets \log\left( \frac{1 - p_j}{p_j} \right)$ \Comment{Item difficulty via inverse of the Rasch probability function}
    \EndFor

    \Statex \textbf{Step 2: Estimate LLM abilities via maximum likelihood using the Rasch model}
    \For{$i = 1$ to $M$}
        \State Define likelihood function:
        \begin{align*}
            \mathcal{L}(\theta_i) = \sum_{j=1}^{I} \:  s_{ij} \cdot log\left ( \frac{1}{1+e^{-(\theta_i - b_j)}} \right ) + \left ( 1 - s_{ij} \right ) \cdot log\left ( 1 - \frac{1}{1+e^{-(\theta_i - b_j)}} \right )
        \end{align*}
        \State Estimate $\theta_i = \arg\max_{\theta} \mathcal{L}(\theta)$ \Comment{MLE for the Rasch model}
    \EndFor

    \Statex \textbf{Step 3: Compute LLM percentile ranks w.r.t. student ability distribution}
    \State Let $\Phi(\theta)$ be the cumulative distribution function (CDF) of student abilities
    \For{$i = 1$ to $M$}
        \State $\pi_i \gets \Phi(\theta_i) \times 100)$ \Comment{Percentile rank of LLM $i$}
    \EndFor

\end{algorithmic}
\end{algorithm*}

\end{document}